\documentclass[twoside]{article}

\usepackage[preprint]{aistats2027}

\usepackage[round]{natbib}

\usepackage[utf8]{inputenc} % allow utf-8 input
\usepackage[T1]{fontenc}    % use 8-bit T1 fonts
\usepackage{hyperref}       % hyperlinks
\usepackage{url}            % simple URL typesetting
\usepackage{booktabs}       % professional-quality tables
\usepackage{amsfonts}       % blackboard math symbols
\usepackage{nicefrac}       % compact symbols for 1/2, etc.
\usepackage{microtype}      % microtypography
\usepackage[dvipsnames]{xcolor}         % colors
\usepackage{amsmath}
\usepackage[ruled,vlined]{algorithm2e} % ruled puts lines top/bottom, vlined adds vertical scope lines
\usepackage{graphicx}
\usepackage{subcaption} % Required for subfigures
\usepackage{xspace}
\usepackage{enumitem}
\setlist[itemize]{leftmargin=1.5em}

\usepackage{makecell}
\usepackage{threeparttable}
\usepackage{capt-of}

\newcommand{\ours}{$L_{ASA}$\xspace}
\newcommand{\vxobs}{\mathbf{x}^\text{obs}\xspace}

\begin{document}

% If the title is very long for camera-ready headings, define a shorter one:
%\runningtitle{Keeping Score: Score-Augmented Loss for Neural Likelihood Surrogates}
%\runningauthor{Shen, Kuusela}

\twocolumn[

\aistatstitle{Keeping Score: Adaptive, Tuning-Free Loss Weighting for Score-Augmented Neural Ratio Estimation}

\aistatsauthor{Alexander Shen \And Mikael Kuusela}

\aistatsaddress{ Carnegie Mellon University \And Carnegie Mellon University } ]

\begin{abstract}
    Neural likelihood surrogates (e.g., Neural Ratio Estimation) for stochastic process models are commonly trained via probabilistic classification on simulated data, which forces a tradeoff between surrogate quality and training costs. For structured models where the exact score $\nabla_\theta \log p(x \mid \theta)$ is available, this information can be incorporated into training by augmenting the cross-entropy loss with a score-matching term. However, the optimal weighting of the two losses is not known a priori, and selecting it by hand requires expensive tuning that undercuts the computational savings. We propose an adaptive, tuning-free algorithm that sets the score loss weights during training based on loss gradients, adding minimal overhead to standard classifier training. We evaluate our approach on case studies involving network dynamics and spatial processes, demonstrating that it improves surrogate quality at a drastically lower computational cost than generating more training data. Notably, in some cases, our approach achieves downstream inference performance equivalent to a 10x increase in training data with less than a 1.1x increase in training time.
\end{abstract}

\section{INTRODUCTION}

Stochastic process models (SPMs), which describe systems that evolve randomly over time or space, are used to describe real-world phenomena across a variety of domains, such as climatology, physics, biology, and finance. Like other parametric models, SPMs describe a relationship between parameters $\theta$ and observed data $x$ via the likelihood function $p(x\mid\theta)$. Traditional likelihood-based inference techniques are challenging for complex SPMs, whose likelihood functions are often analytically intractable or computationally expensive to evaluate.

To address this problem, simulation-based inference (SBI) constructs surrogate models (usually neural networks) for statistical quantities of interest using data simulated from the SPM itself \citep{doi:10.1073/pnas.1912789117, annurev:/content/journals/10.1146/annurev-statistics-112723-034123}. By connecting statistical quantities to minimizers of population loss functions, SBI fights intractability with raw data volume and computational power. For SBI practitioners operating under computational constraints, a natural question emerges: how can we maximize surrogate quality subject to a fixed computational budget?

A commonly used SBI method is NRE-A (from the Neural Ratio Estimation family), which estimates the ratio $p(x\mid\theta)/p(x)$ using the binary cross-entropy (BCE) loss \citep{pmlr-v119-hermans20a}. Vanilla NRE-A has seen usage in high energy physics \citep{atlas_2025}, while NRE-A variants have found broader applications in fields such as astronomy \citep{10.1093/mnras/stae995} and cosmology \citep{10.1093/mnras/stac3215}. For practitioners, the key decision under NRE-A is the amount of training data to use, since this affects both surrogate quality and computational costs (e.g., simulation and neural network training). Other decisions like the dataset prior $\pi(\theta)$ or neural network architecture may arise on a case-by-case basis, but dataset size remains the universal lever between quality and cost in NRE-A.

Recent advances in SBI, however, have opened additional options for the NRE-A approach. Of particular note, the ``Mining Gold'' paper \citep{doi:10.1073/pnas.1915980117} proposes an augmented training procedure for NRE-A using the joint score function $\nabla_\theta \log p(x, z \mid \theta)$ when intermediate outputs $z$ in the data generating process are available (which is often the case for SPMs). The paper demonstrates increased sample efficiency on several SBI case studies when score information is included in training; subsequent work shows the same benefits in other SBI approaches \citep{zeghal2022neuralposteriorestimationdifferentiable, Filipp_2026}.

Sample efficiency is certainly an important consideration for SBI practitioners, but it remains unclear whether incorporating score information into NRE-A training constitutes a \emph{more efficient use of computational resources}. For example, the authors of \citet{doi:10.1073/pnas.1915980117} incorporate joint score information by adding an additional score loss term to the NRE-A BCE loss, but they leave the choice of the score loss weight as an unstructured hyperparameter tuning step. Even outside of score information, other work proposing improvements to NRE-A, such as NRE-B \citep{durkan2020contrastive} and NRE-C \citep{miller2022contrastive}, focuses on sample efficiency and leaves questions of practical computational cost open. A 2x improvement in sample efficiency is not useful if it is computationally cheaper to simply train with 2x more data. How do we efficiently incorporate score information into NRE-A surrogate training? Do practitioners then gain from using score information over simply using more training data?

\subsection{Our focus: NRE-A training costs and analytically tractable SPMs}

To complement previous measurements of sample efficiency, we consider the component of computational cost allocated to training the NRE-A surrogate itself. Despite being an equally necessary and potentially expensive part of the NRE-A pipeline, surrogate training costs have received relatively little attention. We aim to fill this gap in the existing literature, giving practitioners a more complete view of the costs and benefits of using score information in NRE-A.

% For practitioners, the optimal allocation of computational budget between data generation and surrogate training relies on the details of the specific inference task, the types of available simulators and surrogates, and any other auxiliary computations needed for a specific scientific domain. Our focus on training time alone is a prioritization of clarity over scope, providing insight into one part of the SBI pipeline that has not been covered by previous work.

We also focus on SPMs with analytically tractable but computationally expensive likelihood functions $p(x\mid~\theta)$. Surrogate models for these SPMs are vital in scientific pipelines that rely on speed just as much as accuracy \citep{annurev:/content/journals/10.1146/annurev-statistics-112723-034123}. For example, spatial datasets like those used in climate science are far too large for efficient computation of exact likelihoods, which usually requires matrix inversions with $\mathcal{O}(n^3)$ time complexity. While previous work has examined the use of the joint score for intractable SPMs \citep{doi:10.1073/pnas.1915980117}, no previous work, to our knowledge, has yet investigated the use of the exact score $\nabla_\theta \log p(x \mid \theta)$ in NRE-A. While the joint score $\nabla_\theta \log p(x , z \mid \theta)$ is an unbiased estimate of the exact score, it adds noise to the training process that the exact score sidesteps.

\subsection{Our contributions}

We propose a tuning-free algorithm that sets score loss weights from loss gradients during training. In our investigations of three realistic SBI case studies, we find that:
\begin{itemize}
    \item Weights matter; a poorly chosen score loss weight can actually hurt surrogate performance. On the other hand, naive grid search for hyperparameter selection adds training costs that clearly overshadow any corresponding gains in performance. For our case studies, we find that adding the score loss via grid-searched weights is at least 3x less efficient than simply using vanilla NRE-A with more data.
    \item Our proposed adaptive weighting algorithm achieves comparable sample efficiency to grid-searched hyperparameters at a small fraction of the cost of a full grid search. Beyond eliminating the grid search, we further reduce the additional per-minibatch training time cost of using score information from $81.3\%$ to $32.8\%$ (both worst-case figures across case studies).
    \item We show that our approach for integrating score information into NRE-A is \emph{substantially} more training-time efficient than simply using more data. In some cases, score information provides a 10x gain in sample efficiency with less than 1.1x increase in training time.
\end{itemize}

From the training perspective, our approach transforms score information from a costly add-on into a practical, high-value component of the SBI toolkit.

\section{PREVIOUS WORK}

\subsection{SBI and neural ratio estimation for stochastic process models}
\label{sec:nler_for_spm}

SBI provides several neural methods for amortized inference under intractable or computationally expensive likelihoods \citep{doi:10.1073/pnas.1912789117, annurev:/content/journals/10.1146/annurev-statistics-112723-034123}. Quantities such as the likelihood $p(x \mid \theta)$ or posterior $p(\theta \mid x)$ can be directly approximated using density estimation techniques \citep{lueckmann2017flexible, papamakarios2019sequential}. Other SBI techniques estimate an un-normalized likelihood or posterior (such as the likelihood-to-evidence ratio $p(x\mid \theta)/p(x)$), leveraging the fact that most parameter inference is based on \emph{ratios} of probability densities. Such neural ratio estimation (NRE) techniques use a  classification approach to approximate these target ratios. NRE-A is the binary classification approach first proposed in \citet{cranmer2016approximatinglikelihoodratioscalibrated} and implemented for the likelihood-to-evidence ratio $p(x \mid \theta)/p(x)$ in \citet{pmlr-v119-hermans20a}. Subsequent work in NRE extends the approach to multiclass classification via contrastive learning in NRE-B
\citep{durkan2020contrastive} and NRE-C \citep{miller2022contrastive}, while other work explores improvements to NRE-A via balancing \citep{delaunoy2022balanced} and marginal truncation \citep{miller2021truncated}. 

% As mentioned, NRE-A and its variants have seen broad use in scientific applications such as higher energy particle physics \citep{atlas_2025}, astronomy \citep{10.1093/mnras/stae995}, and dark matter cosmology \citep{10.1093/mnras/stac3215}.

Spatial processes are a class of SPMs that has received much attention from SBI researchers. \citet{sbi_for_gps} and \citet{LENZI2023107762} apply amortized SBI methods to parameter estimation for Gaussian processes and max-stable processes. The authors of \citet{WALCHESSEN2024100848} demonstrate that NRE-A surrogates provide substantial benefits for likelihood-based inference tasks on spatial processes. \citet{JMLR:v25:23-1134}, \cite{Sainsbury-Dale02012024},  \citet{sainsbury2025neural}, and \citet{sainsburydale2026neuralparameterestimationincomplete} develop neural Bayes methods for diverse spatial settings, including models with censoring, irregular, or incomplete data. In this work, we similarly investigate two case studies occurring in spatial contexts.

Outside of spatial processes, the authors of \citet{10.1145/3604237.3626881} employ SBI methods for calibration of financial market simulators, while the authors of \citet{10.1371/journal.pcbi.1011406} apply SBI to an SPM governing the formation of synapses in the brain. Much of the previous SBI work for SPMs, however, takes place in a ``black-box'' setting, where no known structure of the SPM is used in surrogate training. Focusing on NRE-A, we show that exploiting one such structural property (the score function) yields more accurate surrogates without additional tuning. Future work could extend these gains to other SBI methods and improve surrogate quality in these domains.

\subsection{Mining Gold: leveraging the joint score to improve NRE-A training}
\label{sec:mining_gold}

The ``Mining Gold'' paper \citep{doi:10.1073/pnas.1915980117} examines a particular class of SPMs that generate observed data $x$ from parameters $\theta$ via a sequence of $m-1$ latent state samples $\{z_j\}$. Here, a sample is generated via $z_1 \sim p_1(z_1\mid \theta) \to z_2 \sim p_2(z_2\mid z_1, \theta) \to ... \to x \sim p_m(x \mid z_{< m}, \theta)$. For tractable $p_1, ..., p_m$, the joint score $\nabla_\theta \log p(x, z_{<m} \mid \theta)$ is also tractable. For a differentiable NRE-A surrogate $h_\phi(x, \theta) \approx \log \frac{p(x \mid \theta)}{p(x)}$  parameterized by $\phi$, the predicted score $\nabla_\theta h_\phi(x, \theta)$ can be incorporated into a score loss $L_{Score}$ and combined with the BCE Loss $L_{BCE}$ for training $h_\phi$:
\begin{align}
\begin{split}
\label{eq:generic_combo_loss}
    \phi^* &= \arg\min_\phi L_{BCE} + \alpha L_{Score}
\end{split}
\end{align}
for some weight $\alpha$. The authors demonstrate sample efficiency gains in a number of conventional SBI tasks. Subsequent work in \citet{zeghal2022neuralposteriorestimationdifferentiable} and \citet{zeghalbenchmark2025} leverages this joint score for increased sample efficiency in neural posterior estimation tasks, while the authors of \citet{Filipp_2026} apply the joint score loss to inference in strong lensing cosmology.

The choice of $\alpha$ in Eq.~\eqref{eq:generic_combo_loss} is left as an open hyperparameter tuning problem in \citet{doi:10.1073/pnas.1915980117}, which hinders the practical use of this approach. That work and follow-ups also do not address the fact that training with $L_{Score}$ along with $L_{BCE}$ increases per-minibatch training time. Our analysis considers this computational cost explicitly, investigating the practical impact of using score information on the surrogate training. %side of the SBI pipeline.

% The theoretical benefits of using score information have indeed been borne out in follow up case studies, but several questions remain unanswered. How does one pick $\alpha$, apart from a brute-force grid search? How expensive is adding the score loss to training? If I can easily generate more training data, is it worth it to use score information at all?

% This prior work clearly demonstrates the benefits of leveraging simulator structure when it is available. However, many important practical details, including hyperparameter tuning and actual training time, have not been adequately explored. A method that achieves 2x sample efficiency is not particularly useful if it results in 4x training time due to additional calculations or hyperparameter tuning. Our work takes these practical considerations into account when designing our improved NLER training procedure.

\section{METHODOLOGY}
\subsection{NRE-A Baseline: Training a surrogate likelihood using binary classification}
\label{sec:classifier_trick}

NRE-A \citep{pmlr-v119-hermans20a} trains a neural network surrogate $h_\phi(x, \theta)$ with weights $\phi$ that takes $(x, \theta)$ as input and outputs an estimate of the likelihood-to-evidence ratio (LER) $p(x\mid\theta)/p(x)$, where $p(x) = \int p(x\mid\theta)\pi(\theta) \: d\theta$. This is done by training a classifier $h_\phi(x, \theta)$ to discriminate between $(x, \theta) \sim p(x \mid \theta) \pi(\theta)$ and $(x, \theta) \sim p(x)\pi(\theta)$. More details on this procedure are given in Appendix \ref{app:classifier_trick}.

We use vanilla NRE-A as our baseline SBI procedure due to its simplicity, popularity and applicability to both frequentist and Bayesian inference. NRE-B \citep{durkan2020contrastive} does not estimate the LER (requiring additional sampling steps for posterior construction), and NRE-C \citep{miller2022contrastive} requires additional tuning to estimate the LER. We leave extensions of NRE-A via truncation \citep{miller2021truncated} and balancing \citep{delaunoy2022balanced} to future work.

\subsection{Our Approach: An adaptive, score augmented loss for NRE-A}

We propose an \textbf{A}daptive \textbf{S}core \textbf{A}ugmented loss \ours that augments the NRE-A BCE loss while addressing the gaps mentioned at the end of Section~\ref{sec:mining_gold}. Our approach operationalizes score information for NRE-A without significant additional training cost.

\subsubsection{Augmenting BCE with Exact Score Loss}

Unlike in \citet{doi:10.1073/pnas.1915980117}, we assume that we have access to the ground truth likelihood $p(x \mid \theta)$ and score function $\nabla_\theta \log p(x \mid \theta)$. We thus propose a score loss based on the exact score $\nabla_\theta \log p(x \mid \theta)$
\begin{align}
\begin{split}
\label{eq:score_loss}
    L_{Score} &= \sum_{i=1}^N \mathbf{1}(y_i =1) \alpha_{i}d(\phi,x_i, \theta_i) \\
    d(\phi,x_i, \theta_i) &= \|\nabla_{\theta} h_\phi(x_i, \theta_i) - \nabla_{\theta}\log p(x_i \mid \theta_i)\|_2^2
\end{split}
\end{align}
where $\alpha_i$ is allowed to vary by observation, and $y_i$ is the NRE-A binary label for data point $i$. The combined training loss becomes $L_{BCE} + L_{Score}$ (the $\alpha$ from Eq.~\eqref{eq:generic_combo_loss} is moved inside $L_{Score}$). The indicator function $\mathbf{1}(y_i =1)$ is used because $y_i = 0$ pairings are continuously resampled in our implementation of NRE-A. Restriction to $y_i = 1$ allows us to amortize evaluations of the score function. Furthermore, most likelihood-based inference procedures (e.g., maximum likelihood estimation, confidence sets) do not rely on the deep tails of the likelihood surface, where the majority of the $y_i = 0$ samples are located.

\subsubsection{Adaptive $\alpha_{i}$ selection for $L_{Score}$}

Naively setting $\alpha_{i} = 1$ is not necessarily optimal because $L_{BCE}$ and $L_{Score}$ exist on fundamentally different scales (see Figure~\ref{fig:overall_timing}). Furthermore, $L_{Score}$ and its gradients depend on the absolute scale of $\theta$ (e.g., scaling all $\theta$ values by $c$ will impact $L_{Score}$, but not $L_{BCE}$). A grid search can yield the optimal $\alpha_i$ values, but this additional training cost can easily nullify sample efficiency gains (see Table~\ref{tab:timing_comparison}).

We instead propose an adaptive algorithm inspired by GradNorm \citep{pmlr-v80-chen18a} for multi-task learning. To achieve scale invariance, we match the norms of loss gradients $\nabla_\phi L_{BCE}$ and $\nabla_\phi L_{Score}$. For a model trained via minibatch gradient descent, let $B(i)$ be the minibatch of training data points that contains $(x_i, \theta_i, y_i)$. We then set $\alpha_{i} = \frac{\|g^{BCE}_i\|_2}{\|g^{Score}_i\|_2}$, where
\begin{align}
\begin{split}
\label{eq:gn}
    g^{BCE}_i &= \nabla_\phi \sum_{j \in B(i)} \text{BCE}(\phi, x_j, \theta_j, y_j), \\
    g^{Score}_i &= \nabla_\phi \sum_{j \in B(i)} \mathbf{1}(y_j=1) d(\phi, x_j, \theta_j),
\end{split}
\end{align}
and $BCE(\phi, x_j, \theta_j, y_j)$ is the BCE loss function evaluated for $h_\phi(x, \theta)$ at data point $j$. In other words, we set $\alpha_{i}$ such that the minibatch gradient of $L_{BCE}$ with respect to model weights $\phi$ has the same magnitude as the gradient of $L_{Score}$. We hold $\alpha_{i}$ constant across all data points in the minibatch to avoid an excessive number of gradient operations.

\subsubsection{Sparse updates for $\alpha_{i}$ via recency-weighted moving average}

For fixed $\alpha_{i}$, the neural network update gradient for \ours can be calculated in a single pass as $\nabla_\phi (L_{BCE} + L_{Score})$. However, our adaptive $\alpha_{i}$ requires calculating both $\|g^{BCE}_i\|_2$ and $\|g^{Score}_i\|_2$ in Eq.~\eqref{eq:gn}, which then needs both $\nabla_\phi L_{BCE}$ and $\nabla_\phi L_{Score}$ separately for \emph{every minibatch}.

To reduce computational cost associated with separate gradient calculations, we replace the continuously updated $\alpha_{i}$ values with a historical moving average. Every $T_\alpha$ minibatches, we compute $\alpha_{i}$ according to Eq.~\eqref{eq:gn} and accumulate an $l$-step history of these values. For the other $T_\alpha-1$ minibatches, we globally set $\alpha_{i}$ equal to a recency-weighted average of the $l$ historical values. We can then use the combined gradient $\nabla_\phi (L_{BCE} + L_{Score})$ for those minibatches. In our experiments, we manually select $T_\alpha=64$ and $l=64$. See Figure~\ref{fig:overall_timing} and Appendix~\ref{app:cost_breakdown} for specific cost savings. 

\subsubsection{Approximation of estimated score using finite differencing}

Calculating $\nabla_\phi L_{Score}$ involves the nested gradient $\nabla_\phi \nabla_\theta h_\phi(x_i, \theta_i)$, which is an expensive operation for deep neural networks. Direct gradients for $L_{Score}$ alone can add around 80\% more per-minibatch training costs in the worst cases (see Figure~\ref{fig:overall_timing}, Appendix~\ref{app:cost_breakdown}). To address this, we approximate $\nabla_\theta h_\phi(x_i, \theta_i)$ using finite differencing
\begin{align*}
    \nabla_{\theta^k} h_\phi(x_i, \theta_i) \approx \frac{h_\phi(x_i, \theta_i + \epsilon_k e_k) - h_\phi(x_i, \theta_i)}{\epsilon_k},
\end{align*}
where $\nabla_{\theta^k}h_\phi(x_i, \theta_i)$ is the $k$'th component of the gradient, $\epsilon_k$ is a small scalar and $e_k \in \mathbb{R}^d$ is the unit vector along dimension $k$. This approximation removes the backpropagation step in calculating $L_{Score}$ while still allowing gradients to propagate for $\nabla_\phi L_{Score}$. This also allows $\nabla_\phi L_{BCE}$ and $\nabla_\phi L_{Score}$ to be computed in the same GPU batch. We tune the perturbation size $\epsilon_k$ by matching a small set of true $\nabla_\theta h_\phi(x_i, \theta_i)$ values at the beginning of training (see Appendix \ref{app:training} for details).

\section{EXPERIMENTS}

We consider three SPM case studies: SIS epidemic models, Gaussian spatial processes, and Student-$t$ spatial processes. Unlike traditional SBI benchmarks (e.g., two moons or Lotka-Volterra), these SPMs are highly relevant to contemporary scientific applications. Usable surrogates for these SPMs are also much larger (thousands to millions of neural network parameters), requiring lengthy training runs on large datasets. These case studies therefore better reflect the scale and cost trade-offs that practitioners actually face.

\begin{figure}[h]
    \centering
    \begin{subfigure}[b]{0.9\linewidth}
        \centering
        \includegraphics[width=\textwidth]{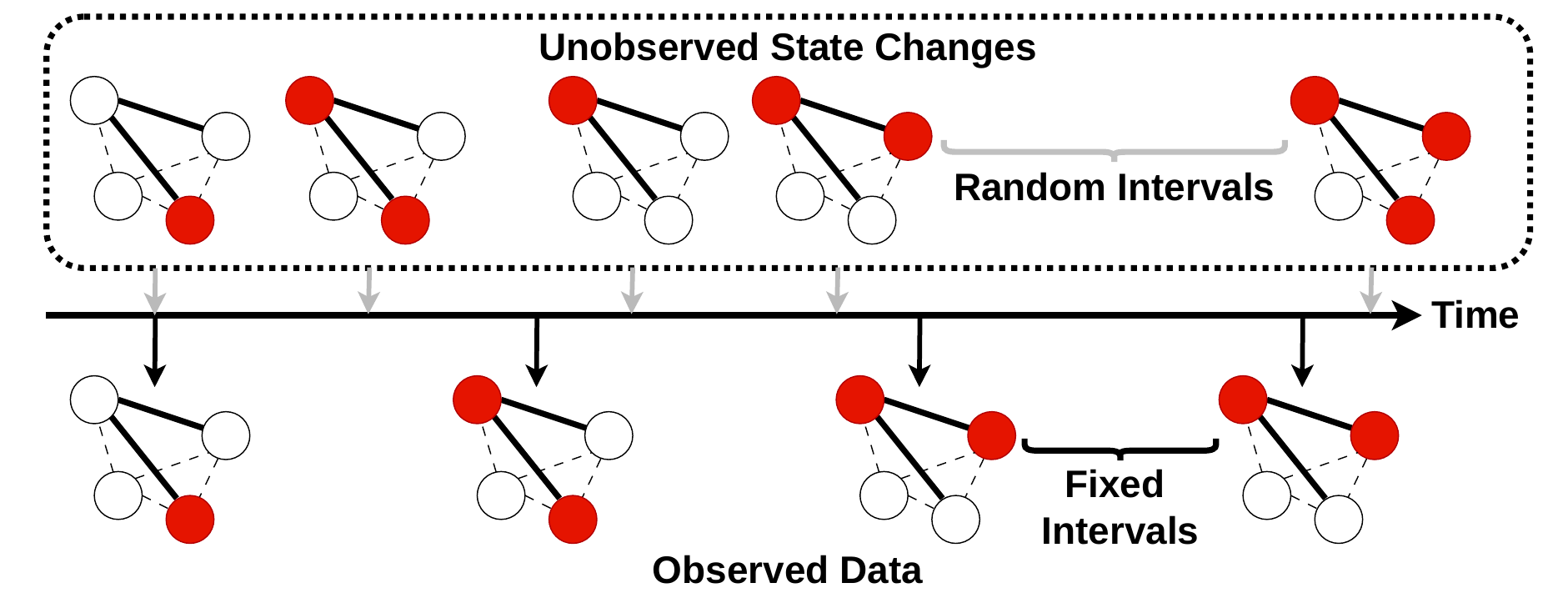}
        \caption{SIS continuous-time hidden Markov model}
    \end{subfigure}
    \vspace{1em}
    \begin{subfigure}[b]{0.9\linewidth}
        \centering
        \includegraphics[width=\textwidth]{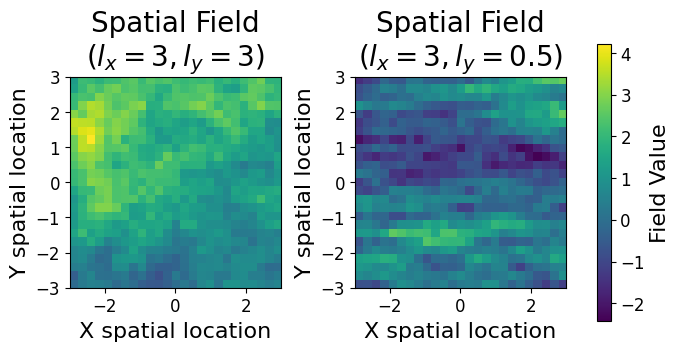}
        \caption{Gaussian/Student-$t$ Process}
    \end{subfigure}
    \caption{\textbf{(Top) Continuous-time hidden Markov model for SIS epidemics}: Four nodes (illustrative), each either infected (red) or susceptible (white), are connected in a graph with differing edge weights (solid vs.\ dashed). The graph undergoes intrinsic, single-node changes at random intervals. Observations are made at fixed intervals, only revealing the graph's current state. \textbf{(Bottom) Gaussian/Student-$t$ spatial process}: Single field realizations of anisotropic spatial processes for different values of the $y$-direction length scale.}
    \label{fig:data}
\end{figure}

Figure~\ref{fig:data} shows example data generated by each SPM. For each case study, we consider several NRE-A scenarios that differ in size of the training dataset $N$ and the size of the neural network surrogate $h_\phi(x, \theta)$. In a given scenario, we examine the performance and training costs of the following NRE-A procedures:
\begin{itemize}
    \item Train using $L_{BCE}$ alone.
    \item Train using $L_{BCE}$ and $L_{Score}$ as in Eq.~\eqref{eq:generic_combo_loss}, choosing $\alpha$ via grid search (denoted GS-$\alpha$).
    \item Train using \ours.
\end{itemize}

See Appendix \ref{app:exp_spec} for surrogate architectures and other experimental hyperparameters in our case studies.

For training cost, we directly measure the time taken to train the likelihood surrogate on a fixed hardware setup (see Appendix~\ref{app:hardware} for details). For simplicity, we measure only those training steps necessary to update weights $\phi$ for the surrogate $h_\phi(x, \theta)$, which excludes validation epochs and auxiliary metric calculations. We further measure time taken to arrive at the best model checkpoint, excluding any additional epochs that occur beyond it (e.g., for convergence patience). In this manner, we remove any effects of training hyperparameters and procedures that do not actually train $h_\phi(x, \theta)$, focusing directly on the optimization process itself.

For performance, we first evaluate $L_{BCE}$ over a large held-out test set for each trained surrogate. Since the true log LER minimizes the population BCE, we use this estimated population BCE to measure how well each surrogate approximates the LER. 

We additionally compute quality metrics related to likelihood inference tasks that better capture surrogate quality from a practitioner's perspective:

\begin{itemize}
    \item Likelihood ratio test statistic (LRTS) distributional discrepancies vs ground truth LRTS
    \item Confidence set empirical coverage and size
    \item Discrepancy between surrogate maximum likelihood estimates (MLEs) of $\theta$  vs ground truth MLEs
\end{itemize}

See Appendix~\ref{app:perf_metrics_def} for details on how each performance metric is calculated.

\subsection{Case study 1: Continuous-time hidden Markov model for SIS Epidemics (SIS CT-HMM)}

The Susceptible-Infected-Susceptible (SIS) epidemic model is used to model disease transmission in a population where individuals without the disease (susceptible) become infected and recover, but can be infected again. Mutating pathogens (e.g., the common cold, influenza, and COVID-19) and their transmission through social networks are amenable to this type of model \citep{app122110934}.

We model a small population using $K=8$ nodes in a weighted graph. Nodes themselves are either in an infected state (1) or susceptible state (0). We assume that intrinsic node transitions are driven by a continuous-time Markov chain (CTMC) parameterized by an (external) infection rate $\lambda$, spontaneous infection rate $\eta$, and recovery rate $\mu$. For this analysis, we set $\eta=0.135$ and perform inference on $\lambda$ and $\mu$. See Appendix~\ref{app:sis_spec} for the exact definition of the CTMC.

We observe graph states $x_{t_i} \in \{1, ..., 2^K\}$ only at certain times $t_i$ independent from any intrinsic transitions. From a sequence of observed states $\{x^\text{obs}_{t_i}\}_{i=1}^T$, the exact likelihood is given by
\begin{align}
\begin{split}
\label{eq:sis_likelihood}
&p( \{x_{t_i}^\text{obs}\}_{i=1}^T\mid\lambda, \mu, \eta ) = \\
&P(x^\text{obs}_{t_1}) \prod_{i=2}^T \left[ \exp \left( Q(\lambda, \mu, \eta) (t_i - t_{i-1}) \right) \right]_{x^\text{obs}_{t_{i-1}}, x^\text{obs}_{t_i}},
\end{split}
\end{align}
where $P(x^\text{obs}_{t_1})$ is the probability of observing $x^\text{obs}_{t_1}$ (some distribution over initial states), $Q(\lambda, \mu, \eta)\in\mathbb{R}^{2^K \times 2^K}$ is the generator matrix corresponding to the parameters, and $[\cdot]_{x^\text{obs}_{t_{i-1}}, x^\text{obs}_{t_i}}$ indicates the row corresponding to state $x^\text{obs}_{t_{i-1}}$ and the column corresponding to state $x^\text{obs}_{t_i}$. $Q$ encodes the instantaneous rates of changing from state $x^\text{obs}_{t_{i-1}}$ to state $x^\text{obs}_{t_i}$ \citep{https://doi.org/10.1111/j.1467-9868.2005.00508.x}; see Appendix \ref{app:sis_spec} for details on how it is constructed.

Importantly, the exponential function in Equation~\eqref{eq:sis_likelihood} is the \emph{matrix exponential}. For CT-HMMs, this operation accounts for all possible intrinsic trajectories between observed states. Most matrix exponential algorithms are at least $\mathcal{O}((2^K)^3)$, making exact likelihood calculations for large numbers of nodes infeasible. See Appendix~\ref{app:inference_time} for timing studies of surrogate likelihood evaluation vs exact likelihood evaluation.

% Figures \ref{fig:sis_1}, \ref{fig:sis_2}, and \ref{fig:sis_3} show test set metrics for NRE-A surrogates in this simulation setting. For this case study, we see improvements for the vast majority of surrogate likelihoods when moving from $L_{BCE}$ to \ours, while paying a minimal additional training cost (see Appendix Figure \ref{fig:sis_timing_all} for additional train time metrics). In particular, the benefits are more pronounced for larger $N$ and larger neural network sizes, which is the preferred regime for SBI methods.

% \begin{figure}[ht]
%     \centering
%     \includegraphics[width=\columnwidth]{figures/sis_mle_mse_surfaces.png}

%     \caption{\textbf{CT-HMM for SIS epidemics maximum likelihood estimator (MLE) metrics and example likelihood surfaces: (Top)} MSE of surrogate MLE with respect to true MLE for same $\mathbf{x_{i,g}}$. Mean $\pm2$ standard errors over 5 seeds is plotted for each $h_\phi(x, \theta)$ size (vertical lanes), training dataset size (marker colors) and loss function (marker shape). \ours results in improved or unaffected MLE performance across most NRE-A settings. \textbf{(Bottom)} Example surrogate likelihood surfaces for same $x^\text{obs}$, scaled relative to maximum and with associated contours. Both surfaces use $h_\phi(x, \theta)$ size 100k and 100k training data points. We see that \ours modifies the shape of the entire surrogate likelihood surface, ultimately shifting the surrogate MLE closer to the ground truth MLE (white diamond).}
%     \label{fig:sis_3}
% \end{figure}

\subsection{Case study 2: Anisotropic Gaussian spatial process with nugget effect (GP)}

Stochastic process models for spatial data have been used to describe diverse physical phenomena, such as weather patterns, astronomical surveys, and oceanographic fields \citep{gelfand2010handbook}. In particular, the Gaussian process (GP) model provides a simple but flexible baseline for understanding spatial dynamics in physical systems.

In our case study, we consider a zero-mean GP with an anisotropic exponential covariance over 2D space. In addition to two length scales $(l_x, l_y)$, we also add a small nugget variance parameter $\varepsilon^2$ to model small-scale variations that are independent across spatial locations. We also fix a $25 \times 25$ grid of sample locations over the 2D spatial area $[-3, 3] \times [-3, 3]$. Let $\mathbf{x}^\text{obs} \in \mathbb{R}^S$ be a sampled field at these $S=625$ locations and let $\Sigma(l_x, l_y, \varepsilon)$ be the covariance matrix generated by $(l_x, l_y, \varepsilon)$. The likelihood given $\mathbf{x}^\text{obs}$ is defined as
\begin{align}
\begin{split}
\label{eq:gp_likelihood}
p(\vxobs \mid l_x, l_y, \varepsilon) &= (2\pi)^{-S/2} |\Sigma(l_x, l_y, \varepsilon)|^{-1/2} \times\\
&\exp\left( -\frac{1}{2} (\vxobs)^\top \Sigma(l_x, l_y, \varepsilon)^{-1} \vxobs \right).
\end{split}
\end{align}
See Appendix \ref{app:gn_spec} for additional details on how the GP covariance is defined and Figure \ref{fig:data} for examples of sampled data. While the likelihood is analytically tractable, it is computationally expensive for large values of $S$ due to the matrix inversion and determinant. See Appendix~\ref{app:inference_time} for timing studies.

% Figure \ref{fig:gn_1} shows the L-test set metrics for this experimental setting. As with the previous case study, we see broad improvements in surrogate performance without a significant increase in training cost. For E-test metrics and additional training time metrics on this case study, see Figures \ref{fig:gn_2}, \ref{fig:gn_3}, and \ref{fig:gn_timing_all} in Appendix \ref{app:gn_results}.

\subsection{Case study 3: Anisotropic Student-$t$ spatial process (STP)}

The Student-$t$ process (STP) \citep{pmlr-v33-shah14} is a generalization of the Gaussian process with heavier tails. Much like the univariate Student-$t$ distribution, the mean-zero STP can be generated by sampling a mean-zero Gaussian process and using a $\chi^2_\nu$ random sample to scale the field. Examples of spatial fields from a Student-$t$ process are shown in Figure \ref{fig:data}.

Let $\Sigma(l_x, l_y, \varepsilon)$ be the exponential covariance corresponding to length scales $(l_x, l_y)$ and nugget effect $\varepsilon$. Let $\mathbf{x}^\text{obs} \in \mathbb{R}^S$ be an STP sampled at locations $\mathbf{s}$. The likelihood given observed data $\vxobs$ is defined as

\begin{align}
\begin{split}
\label{eq:stp}
    p(\vxobs \mid \theta) &= \frac{\Gamma(\frac{\nu + S}{2})}{\pi^{S/2}(\nu-2)^{S/2}\Gamma(\nu/2)}|\Sigma(l_x, l_y, \varepsilon)|^{-1/2}\times \\
    &\left(1+\frac{(\vxobs)^\top\Sigma(l_x, l_y,\varepsilon)^{-1}\mathbf{x}^\text{obs}}{\nu-2}\right)^{-\frac{\nu+S}{2}},
\end{split}
\end{align}
where $\theta = (l_x, l_y, \varepsilon, \nu)$. We fix $\varepsilon=0$ for the STP case study and perform likelihood estimation for the two length scales $(l_x, l_y)$ and degrees of freedom $\nu$. As with the GP, exact likelihood evaluation is computationally constrained by the matrix inversion and determinant (see Appendix~\ref{app:inference_time} for our timing study).  

% Figure \ref{fig:stp_1} shows L-test set metrics for the Student-$t$ process. As with the previous case study, we see broad improvements in surrogate performance without a significant increase in training cost. For E-test metrics and additional training time metrics on this case study, see Figures \ref{fig:stp_2}, \ref{fig:stp_3}, and \ref{fig:stp_timing_all} in Appendix~\ref{app:stp_results}.

\section{RESULTS}
\label{sec:results}
For notational purposes, let $h^*_\phi(x, \theta)$ be the surrogate architecture that achieves the lowest test BCE at $N=1$m for each case study (size 100k, 100k, and 30k for the SIS CT-HMM, Gaussian process, and Student-$t$ process case studies respectively). Unless otherwise specified, all error bars/bands represent $\pm 2$ standard errors of the mean across 5 independent seeds.

\subsection{\ours solves previous inefficiencies in incorporating score information}

We find that \ours 
increases NRE-A surrogate quality at a lower additional training cost than either GS-$\alpha$ or simply adding more training data. Table~\ref{tab:timing_comparison} shows that for larger dataset scenarios, \ours can achieve surrogate quality improvements associated with 3x more training data \emph{while simultaneously reducing total training time.} Both $L_{BCE}$ and GS-$\alpha$ require \emph{more} training time to achieve the same surrogate improvement. 

Figure~\ref{fig:overall_timing} shows how each component of \ours contributes to improved training efficiency. We confirm that the choice of $\alpha$ matters: a poorly chosen $\alpha$ can actually \emph{hurt} surrogate performance. In contrast, \ours closely matches the performance of GS-$\alpha$ while skipping the grid search and adding only 17.2\%--32.8\% more per-minibatch training time in comparison to $L_{BCE}$.

\begin{table}[!ht]
\centering
\begin{threeparttable}
\caption{Costs of gains over $L_{BCE}$ at $N=300$k}
\label{tab:timing_comparison}
\begin{tabular}{lccc}
\toprule
 & SIS & GP & STP \\
\midrule
\makecell[l]{$L_{BCE}$ (1m)} & $3.53 \pm 1.1$ & $1.63 \pm 0.6$ & $1.88 \pm 0.6$ \\
\addlinespace[1pt]
\makecell[l]{GS-$\alpha$} & $12.18 \pm 7.9$ & $6.89 \pm 1.8$ & $9.16 \pm 4.5$ \\
\addlinespace[1pt]
\makecell[l]{\ours} & $\mathbf{0.66 \pm 0.4}$ & $\mathbf{0.39 \pm 0.1}$ & $\mathbf{0.54 \pm 0.3}$ \\
\bottomrule
\end{tabular}
\begin{tablenotes}[flushleft]
\small
\item For each case study (columns), let the ``base cost'' be the training time for $h^*_\phi(x, \theta)$ and $L_{BCE}$ alone at $N=300$k. We define the ``target performance gain'' as the decrease in test BCE training with $L_{BCE}$ alone at $N=1$m instead of $N=300$k. Each cell reports the training time until the \emph{first} model checkpoint that achieves the target performance gain, relative to the base cost (e.g., a value of $2$ indicates 2x base cost). Rows indicate which procedure is used to gain performance ($N=300$k unless specified). GS-$\alpha$ reports best $\alpha$ time multiplied by 13 grid search points.
\end{tablenotes}
\end{threeparttable}
\vspace{0.5em}
\hrule
\vspace{0.5em}

\includegraphics[width=\linewidth]{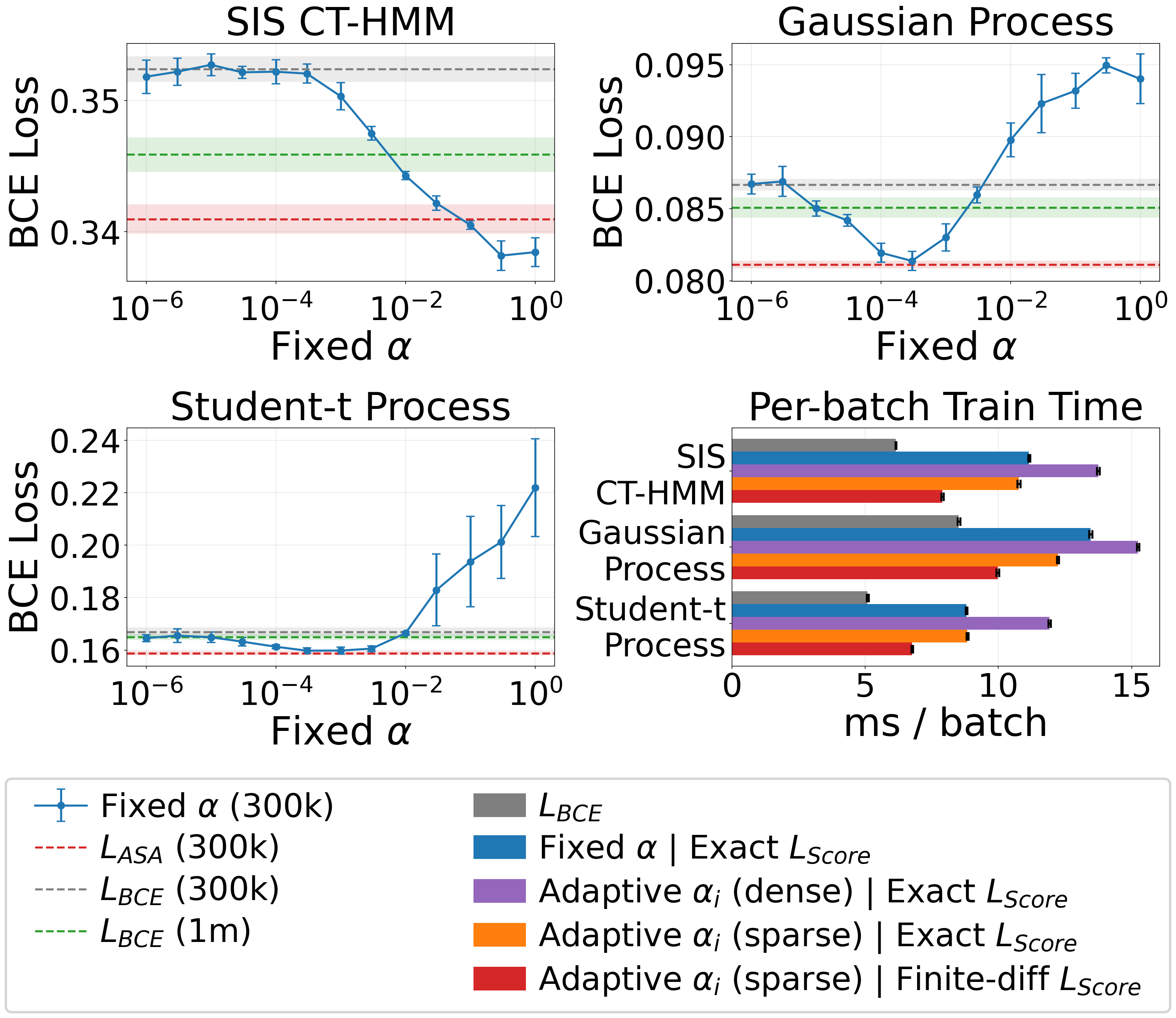}
\captionof{figure}{\textbf{(Top Left, Top Right, Bottom Left)} For each fixed $\alpha$ for $L_{Score}$, we plot the test BCE for $h^*_\phi(x, \theta)$ and $N=300$k (blue dots). The wrong $\alpha$ can worsen original NRE-A performance (gray), while a suitable $\alpha$ can achieve equivalent $L_{BCE}$-alone performance at $N=1$m (green). \ours (red) achieves performance nearly matching the full grid search. \textbf{(Bottom Right)} Per-minibatch computation time of different training procedures, averaged over $\sim$9,000 batches. We compare $L_{BCE}$ alone (gray) with fixed $\alpha$, exact $L_{Score}$ via autograd for $\nabla_\theta h_\phi(x, \theta)$ (blue), using adaptive $\alpha_i$ with weights updated every minibatch (purple), using our sparse $\alpha_i$ updates  (orange), and using finite-difference for $L_{Score}$ (red). While baseline $L_{Score}$ increases per-minibatch training time by 58.1\%--81.3\%, \ours only requires 17.2\%--32.8\% more time.}
\label{fig:overall_timing}
\hrule
\end{table}

% \begin{figure}[h]
%     \centering
%     \includegraphics[width=1\linewidth]{figures/overall_timing.png}
%     \caption{\textbf{(Top Left, Top Right, Bottom Left)} For each fixed $\alpha$ for $L_{Score}$, we plot the corresponding test BCE for $h^*_\phi(x, \theta)$ and $N=300$k (blue dots). We find that the wrong $\alpha$ can worsen original NRE-A performance (gray), while a suitable $\alpha$ can achieve equivalent $L_{BCE}$-alone performance at $N=1$m (green). \ours (red) achieves performance nearly matching the full grid search. \textbf{(Bottom Right)} Per-minibatch computation time of different training procedures, averaged over $\sim$9000 batches. We compare $L_{BCE}$ alone (gray) with fixed $\alpha$, exact $L_{Score}$ via autograd for $\nabla_\theta h_\phi(x, \theta)$ (blue), using adaptive $\alpha_i$ with weights updated every minibatch (purple), using our sparse update procedure for $\alpha_i$ (orange), and using finite-difference for $L_{Score}$ (red). While baseline $L_{Score}$ increases per minibatch training time by 58.1\%-81.3\%, \ours only requires 17.2\%-32.8\% more time. }
%     \label{fig:overall_timing}
% \end{figure}

\subsection{Using \ours substantially boosts surrogate quality with low additional training cost }

\begin{figure}[!ht]
    \centering
    \hrule
    \vspace{0.5em}
    \includegraphics[width=\columnwidth]{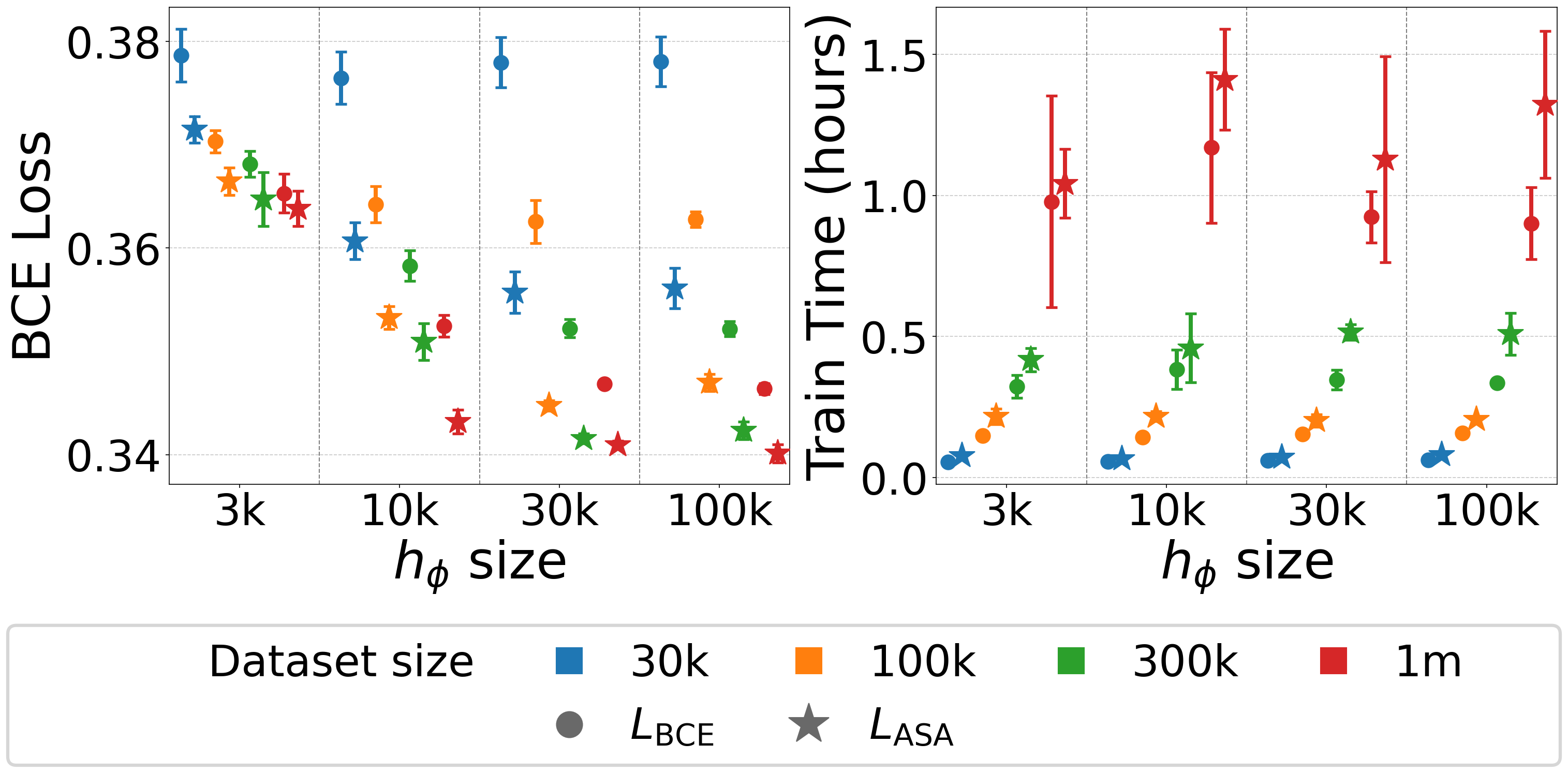}

    \vspace{1em}
    \includegraphics[width=\columnwidth]{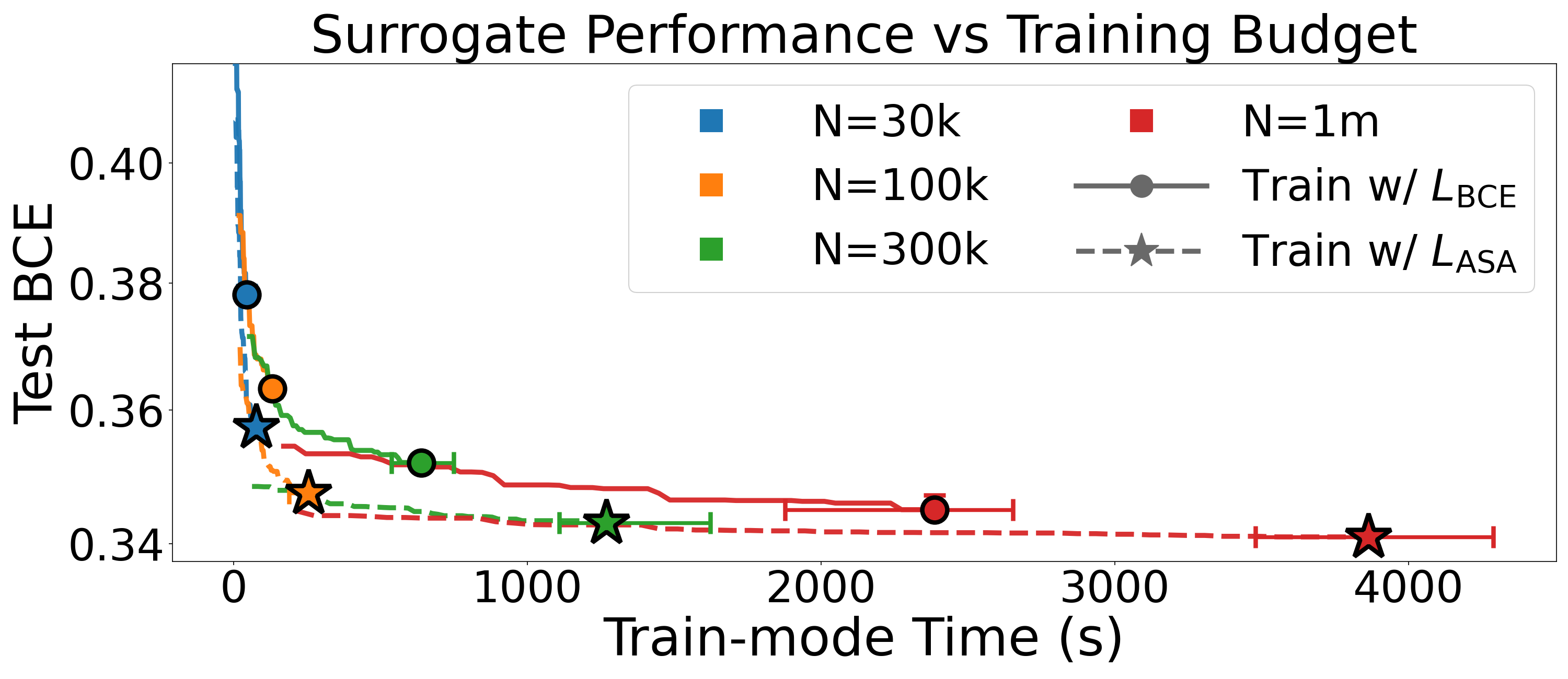}
    \caption{\textbf{CT-HMM for SIS epidemics: (Top Left)} For each $h_\phi(x, \theta)$ size (vertical lanes) and training dataset size (marker colors), we compute test BCE   for the surrogate trained under $L_{BCE}$ (dot) vs.\ trained under $L_{ASA}$ (star). Especially for larger model sizes and larger training datasets, using \ours can be equivalent to training with 3x or 10x more training data under $L_{BCE}$. \textbf{(Top Right)} Same plot for total training time. Gains in surrogate performance via \ours cost much less than simply increasing dataset size. \textbf{(Bottom)} For $h^*_\phi(x, \theta)$, we plot the median (5 seeds) test BCE over the course of training. The final point is plotted at the median of total training time (error bars show IQR). For the same training time, \ours achieves substantially better surrogate performance than $L_{BCE}$ alone. }
    \label{fig:sis_1}
    \hrule
\end{figure}

\begin{figure}[!ht]
    \centering
    \includegraphics[width=\columnwidth]{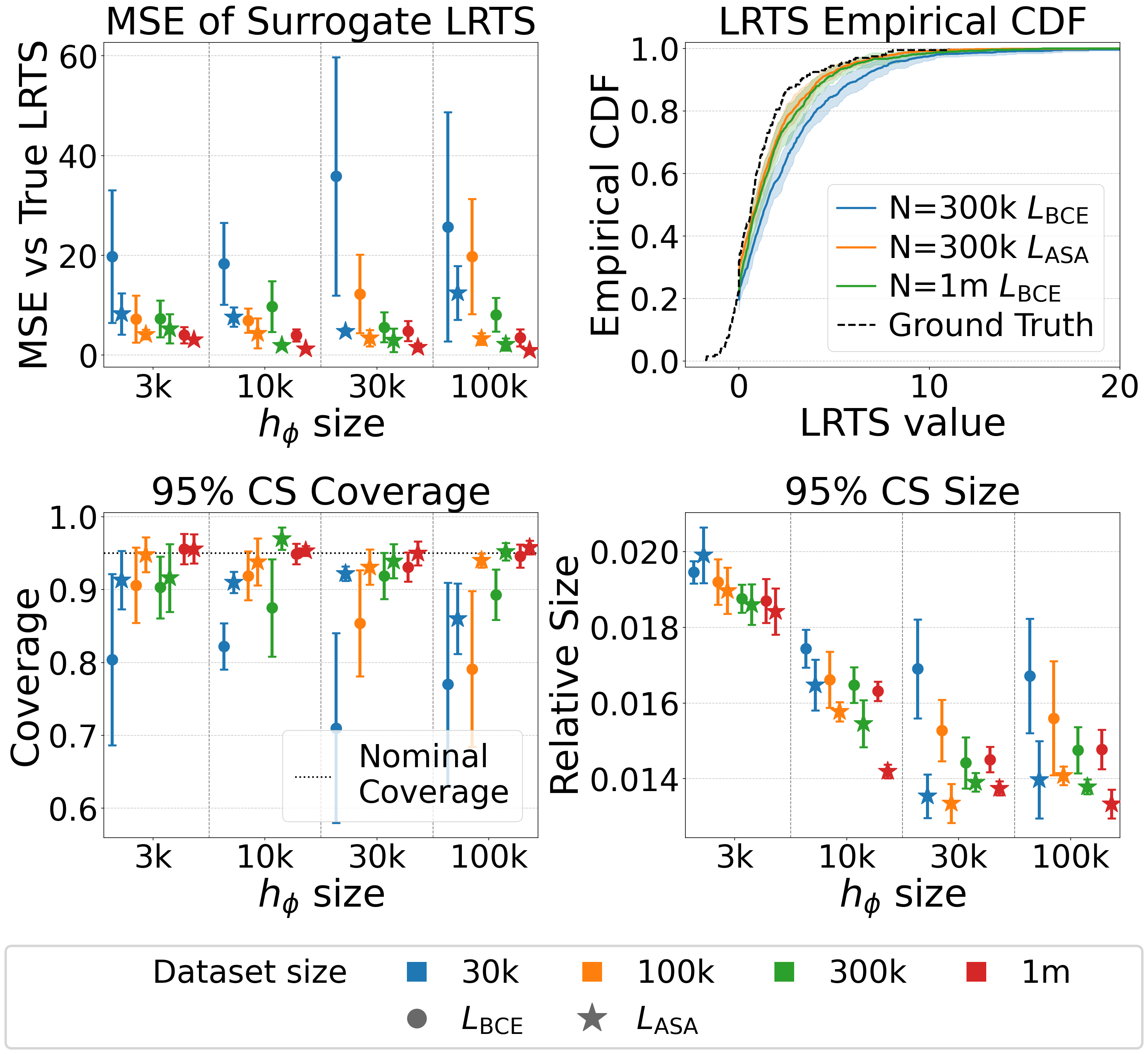}
    
    \caption{\textbf{(Top Left)} MSE between surrogate likelihood ratio test statistic (LRTS) and true LRTS for same $x$, plotted for each $h_\phi(x, \theta)$ size (vertical lanes), training dataset size (marker colors), and procedure (marker shape). Seed noise dominates some settings, but the overall trend favors \ours. \textbf{(Top Right)} Empirical CDF of the LRTS for $h^*_\phi(x, \theta)$. \ours pushes the surrogate CDF closer to ground truth, as if 3x more training data were used. \textbf{(Bottom Left)} Empirical coverage of 95\% Wilks' confidence sets. We see that \ours alleviates severe confidence set undercoverage present in some $L_{BCE}$ baselines. \textbf{(Bottom Right)} Average size of 95\% Wilks' confidence sets as a proportion of the parameter space. In many settings, \ours restores coverage while \emph{shrinking} confidence set size. }
    \label{fig:sis_2}
    \hrule
\end{figure}

Because \ours achieves comparable performance to GS-$\alpha$ in an order of magnitude less computing time, we compare surrogate performance between $L_{BCE}$ and \ours only. Figures \ref{fig:sis_1} and \ref{fig:sis_2} show performance metrics for the SIS CT-HMM case study. We see improvements for the vast majority of NRE-A settings when moving from $L_{BCE}$ to \ours, while paying a minimal additional training cost. In particular, the benefits are more pronounced for larger $N$ and larger neural network sizes, which is the preferred regime for SBI methods.

\begin{figure}[!ht]
    \centering
    \includegraphics[width=\columnwidth]{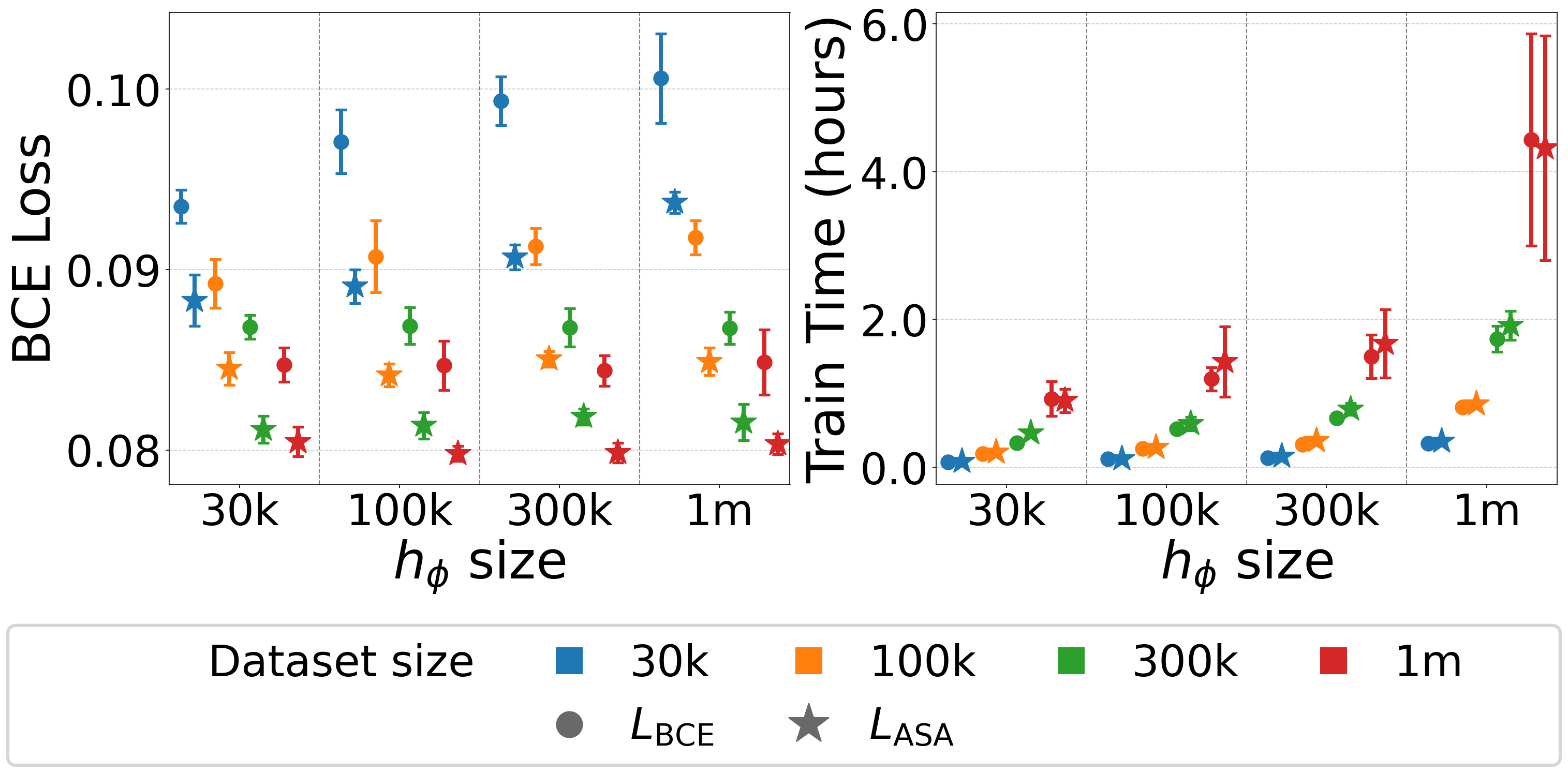}
    \includegraphics[width=\columnwidth]{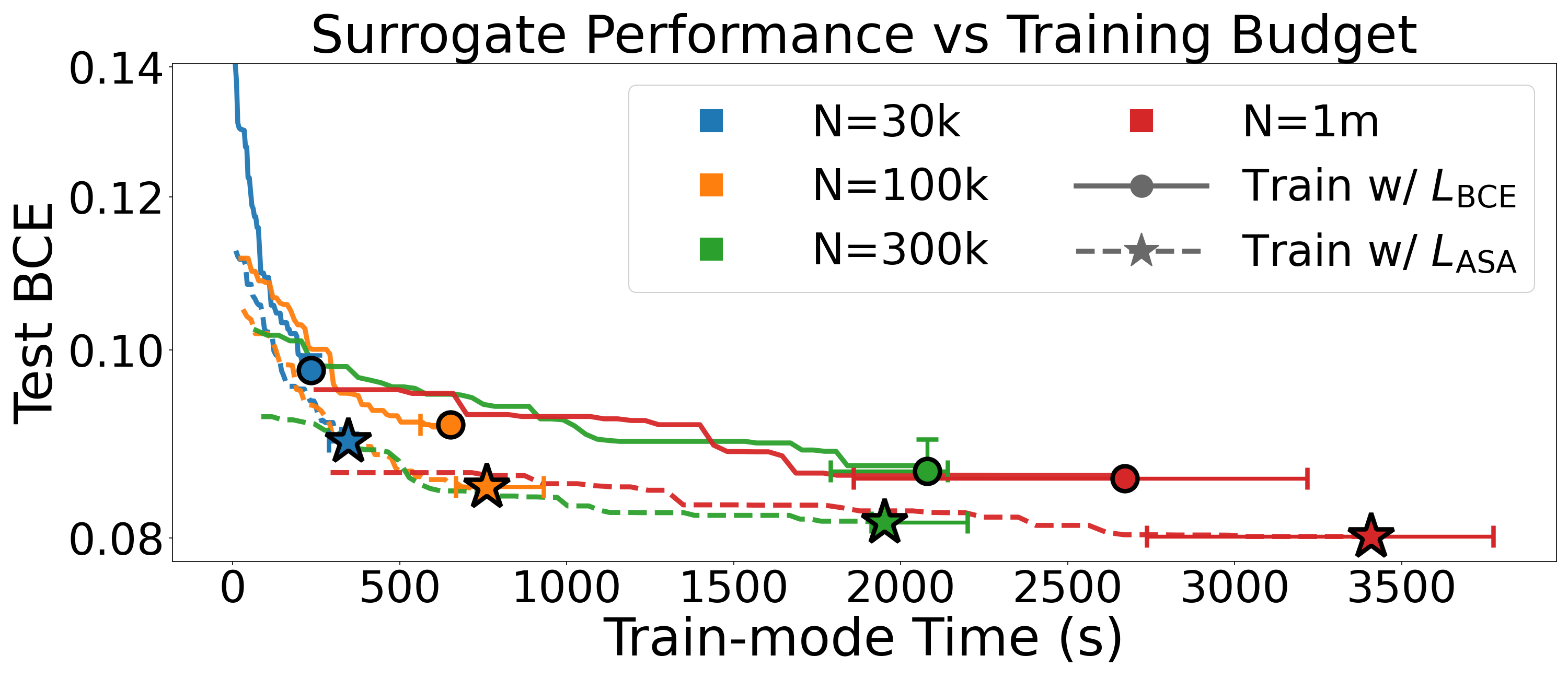}
    \caption{\textbf{Gaussian spatial process}: Figure is formatted identically to Figure~\ref{fig:sis_1}. Although spatial fields are quite different from epidemic network states, we still find that \ours achieves the same performance as $L_{BCE}$ alone with lower training cost. }
    \label{fig:gn_1}
    \hrule
\end{figure}

\begin{figure}[!ht]
    \centering
    \includegraphics[width=\columnwidth]{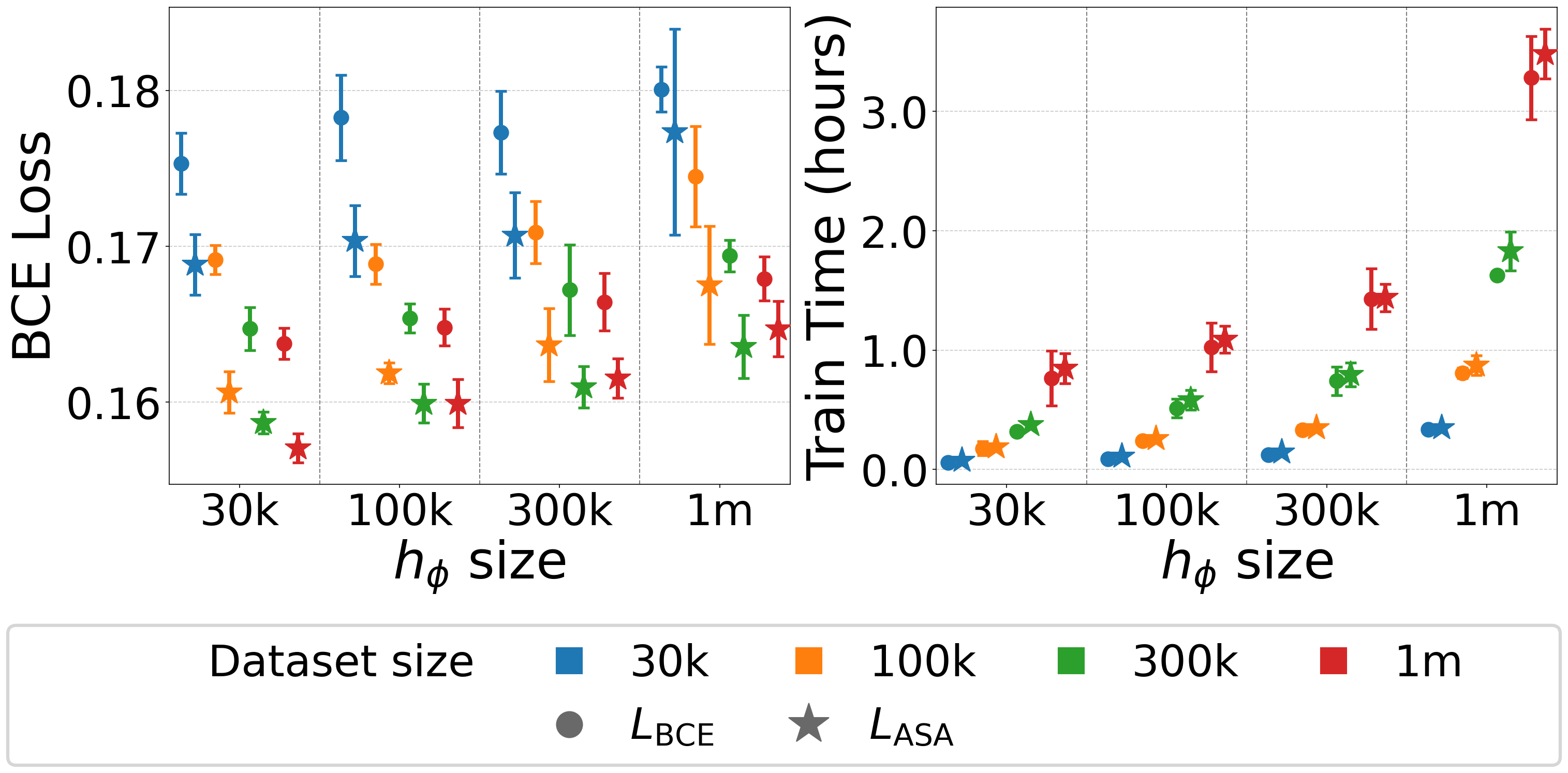}
    \includegraphics[width=\columnwidth]{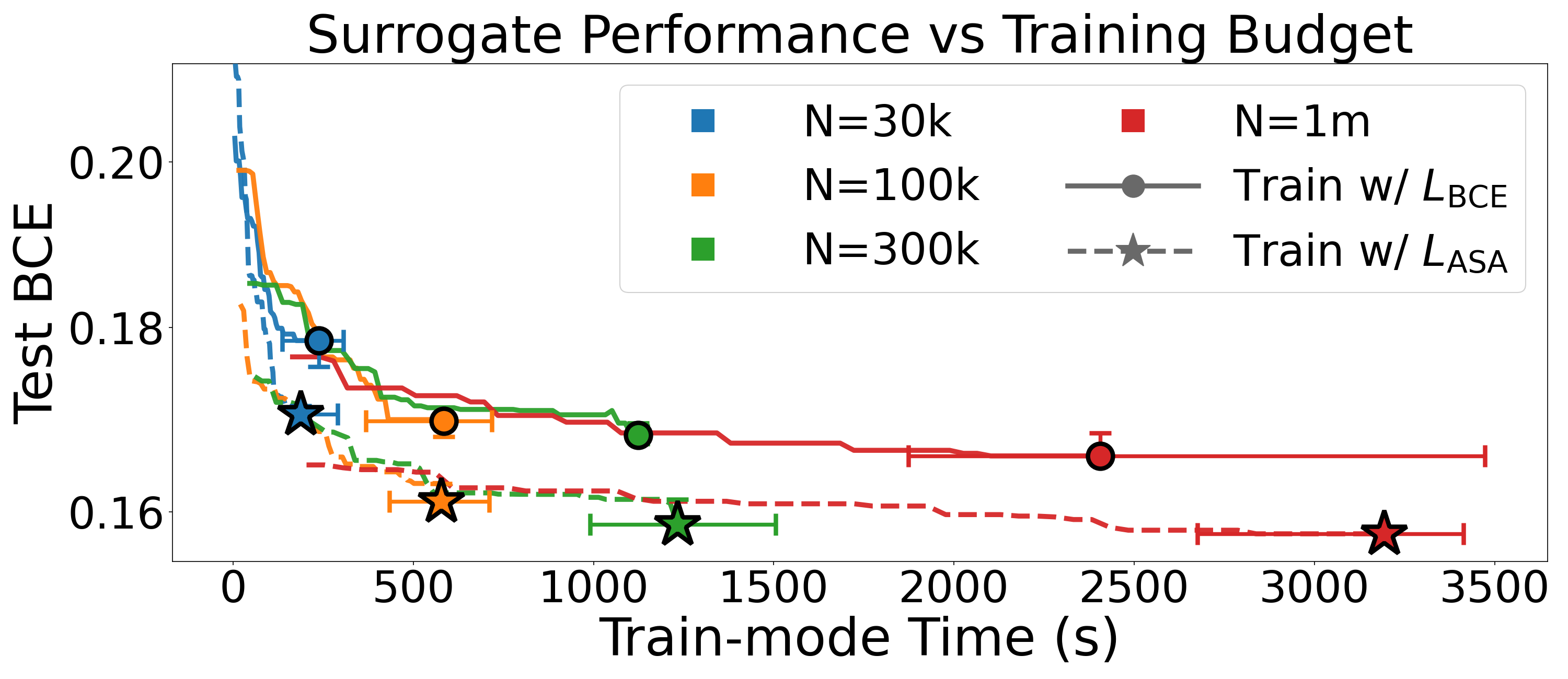}
     \caption{\textbf{Anisotropic Student-$t$ spatial process}: Figure is formatted identically to Figure~\ref{fig:sis_1}. Although the Student-$t$ process has heavier tails vs. its Gaussian counterpart, \ours still delivers significant performance gains with a marginal increase in training cost.}
    \label{fig:stp_1}
    \hrule
\end{figure}

Figures \ref{fig:gn_1} and \ref{fig:stp_1} show similar gains in the other two case studies. We find evidence that incorporating score information via \ours can improve NRE-A efficiency over a broad range of domains and specific training setups. See Appendix~\ref{app:sis_results}, \ref{app:gn_results}, and \ref{app:stp_results} for additional performance metrics.

\section{DISCUSSION}

Our results provide strong additional support for incorporating score information into NRE-A training. While previous work demonstrates better sample efficiency, we show that an additional score loss (with appropriate optimizations) also benefits practitioners in terms of surrogate training costs. \ours also fills the implementation gap of previous work, both by removing the need for an expensive grid search for the score loss weight $\alpha$ and by speeding up the calculation of the score loss itself. In three scientifically relevant, large-scale case studies, we demonstrate how \ours significantly increases surrogate performance without a corresponding jump in training cost.

Considerable avenues for future research remain for incorporation of score information into NRE-A training and for other SBI methods at large. Our method requires an analytically tractable score function, which is not available in some SPMs of interest. Extensions of our work along the lines of \citet{doi:10.1073/pnas.1915980117} could provide new methodologies in those settings. While we investigated different surrogate sizes and training dataset sizes, there are many other factors that could affect final surrogate performance. These include other architecture choices, regularization procedures, and learning rate schedules. Finally, our work here focuses only on the NRE-A approach. Both NRE-B and NRE-C could feasibly benefit from score information in a similar manner, as well as more recent extensions of NRE-A. Many SPMs also incorporate other known structure (e.g., translational and/or rotational invariance in spatial processes) that could potentially be used to augment SBI training procedures.

\clearpage
\subsection*{AI use statement}

% NOTE: This statement is required by AISTATS and does not count against
% the page limit. Please fill in with your actual disclosure before
% submission -- the text below is a placeholder carried over from the
% template and should be edited to reflect what was actually done.

In this work, we used generative AI tools for implementing methods. We have not used generative AI tools for any of the following tasks
\begin{itemize}
    \item generate synthetic data sets
    \item help develop theoretical models or conceptual frameworks
    \item formulate mathematical claims
    \item propose or refine hypotheses
    \item design or provide feedback on research  methodology or experiments
    \item clean and reformat dataset
    \item support qualitative and thematic data analysis
    \item interpret results 
\end{itemize}

and the following are not applicable
to this work:
\begin{itemize}
    \item assist with translation
    \item provide critical ingredients for proving mathematical claims
    \item assist in the writing of proofs
\end{itemize}
Additionally, we used generative AI tools for the following
\begin{itemize}
    \item create or modify scientific figures or images
    \item create or edit software code
    \item creation of artifacts
    \item summarize or analyse existing literature
    \item edit a research paper to improve readability
\end{itemize}

We have reviewed all AI-assisted work. Design of algorithms was done by hand. Code implementation was done mostly by hand with some assistance by generative AI. Correctness was verified through human inspection of intermediate and final outputs by both authors, the same procedure for human written code. Figures and artifacts were checked against raw outputs for consistency. All paper text began as human-written drafts, and all AI-assisted edits passed through human review. All AI claims about existing literature were verified against original texts. We take responsibility for the final content of this work, including text, claims, or artifacts produced with the aid of generative AI.

% \subsubsection*{Acknowledgements}
% Per the AISTATS instructions, acknowledgements should be included only
% in the camera-ready version to preserve anonymity during review.
\bibliographystyle{plainnat}
\bibliography{references}

%%%%%%%%%%%%%%%%%%%%%%%%%%%%%%%%%%%%%%%%%%%%%%%%%%%%%%%%%%%%
\section*{Checklist}

\begin{enumerate}

  \item For all models and algorithms presented, check if you include:
  \begin{enumerate}
    \item A clear description of the mathematical setting, assumptions, algorithm, and/or model. [Yes, see main text and Appendix~\ref{app:exp_spec}]
    \item An analysis of the properties and complexity (time, space, sample size) of any algorithm. [Yes, see main text and Appendix~\ref{app:additional_results}]
    \item (Optional) Anonymized source code, with specification of all dependencies, including external libraries. [No, but it is available upon request]
  \end{enumerate}

  \item For any theoretical claim, check if you include:
  \begin{enumerate}
    \item Statements of the full set of assumptions of all theoretical results. [Not Applicable, no theoretical results]
    \item Complete proofs of all theoretical results. [Not Applicable, no theoretical results]
    \item Clear explanations of any assumptions. [Not Applicable, no theoretical results]
  \end{enumerate}

  \item For all figures and tables that present empirical results, check if you include:
  \begin{enumerate}
    \item The code, data, and instructions needed to reproduce the main experimental results (either in the supplemental material or as a URL). [No, all necessary instructions for reproducing results are included in main text and Appendix~\ref{app:exp_spec}, but code is not provided (it is available upon request). Data generation instructions are clearly provided]
    \item All the training details (e.g., data splits, hyperparameters, how they were chosen). [Yes, see main text and Appendix~\ref{app:exp_spec}]
    \item A clear definition of the specific measure or statistics and error bars (e.g., with respect to the random seed after running experiments multiple times). [Yes, see main text, Section~\ref{sec:results}]
    \item A description of the computing infrastructure used. (e.g., type of GPUs, internal cluster, or cloud provider). [Yes, see Appendix~\ref{app:hardware}]
  \end{enumerate}

  \item If you are using existing assets (e.g., code, data, models) or curating/releasing new assets, check if you include:
  \begin{enumerate}
    \item Citations of the creator If your work uses existing assets. [Yes, see Appendix~\ref{app:hardware}]
    \item The license information of the assets, if applicable. [Yes, see Appendix~\ref{app:hardware}]
    \item New assets either in the supplemental material or as a URL, if applicable. [Not Applicable, no new assets created.]
    \item Information about consent from data providers/curators. [Not Applicable, all data used in analysis is simulated]
    \item Discussion of sensible content if applicable, e.g., personally identifiable information or offensive content. [Not Applicable, all data used in analysis is simulated]
  \end{enumerate}

  \item If you used crowdsourcing or conducted research with human subjects, check if you include:
  \begin{enumerate}
    \item The full text of instructions given to participants and screenshots. [Not Applicable, no crowdsourcing or human subject research]
    \item Descriptions of potential participant risks, with links to Institutional Review Board (IRB) approvals if applicable. [Not Applicable, no crowdsourcing or human subject research]
    \item The estimated hourly wage paid to participants and the total amount spent on participant compensation. [Not Applicable, no crowdsourcing or human subject research]
  \end{enumerate}

\end{enumerate}

\clearpage
\appendix
\thispagestyle{empty}
\onecolumn
\aistatstitle{Keeping Score: Adaptive, Tuning-Free Loss Weighting for Score-Augmented Neural Ratio Estimation: \\
Supplementary Materials}

\section{NRE-A Surrogate Training Procedure}
\label{app:classifier_trick}

NRE-A \citep{pmlr-v119-hermans20a} trains a neural network $h_\phi(x, \theta)$ to approximate the likelihood-to-evidence ratio (LER) $\frac{p(x \mid \theta)}{p(x)}$  where $p(x) = \int p(x\mid\theta)\pi(\theta) \: d\theta$ for some prior $\pi(\theta)$. The choice $\pi(\theta)$ is arbitrary and is often set equal to the design distribution $\pi'(\theta)$ used to generate the training data (e.g., a uniform distribution over some bounded parameter space).

The binary classification approach for NRE-A training starts with generating a training dataset of size $N$ in the following manner:

\begin{align*}
    y_i &\sim \text{Bernoulli}(p) \\
    \text{If } y_i = 1 &\to \theta_i \sim \pi'(\theta), x_i \sim p(x \mid \theta_i) \\
    \text{If } y_i = 0 &\to \theta_i \sim \pi'(\theta), x_i \sim p(x)
\end{align*}

This gives us tuples $(y_i, \theta_i, x_i)$ that form our training data. One could directly use the sequential procedure above, but in practice the dataset is often generated by sampling all $y_i = 1$ tuples first, then forming $y_i = 0$ tuples by permuting the indices on $\theta_i$ and/or $x_i$ such that the dependence between $\theta_i$ and $x_i$ is broken. In principle, one could use any distribution $g(x)$ for sampling $x$ in the $y_i=0$ class, but only the marginal $p(x)$ makes the minimizer of the population BCE loss equal to the LER.

Let $h_\phi(x, \theta)$ be a binary classifier that outputs unnormalized scores such that $\sigma(h_\phi(x, \theta)) = \hat{ \mathbb{P}}(Y = 1 \mid x, \theta)$ for the sigmoid function $\sigma$. We train this classifier using the standard binary cross-entropy (BCE) loss whose empirical form is given below.

\begin{align*}
    L_{BCE} = -\sum_{i=1}^N [y_i \log \sigma(h_\phi(x_i, \theta_i)) + (1 - y_i) \log (1 - \sigma(h_\phi(x_i, \theta_i)))]
\end{align*}

In the limit of infinite data, the BCE loss is minimized by the true conditional probability $\mathbb{P}(Y_i = 1 \mid x_i, \theta_i)$, which leads to the following

\begin{align*}
    \mathbb{P}(Y_i = 1 \mid x_i, \theta_i) &= \frac{p(x_i, \theta_i \mid Y_i = 1)\mathbb{P}(Y_i=1)}{p(x_i, \theta_i)} = \frac{p(x_i \mid \theta_i) \pi(\theta_i)\mathbb{P}(Y_i=1)}{p(x_i, \theta_i)} \\
    \mathbb{P}(Y_i = 0 \mid x_i, \theta_i) &= \frac{p(x_i, \theta_i \mid Y_i = 0)\mathbb{P}(Y_i=0)}{p(x_i, \theta_i)} = \frac{p(x_i) \pi(\theta_i)\mathbb{P}(Y_i=0)}{p(x_i, \theta_i)} \\
    \log \frac{\mathbb{P}(Y_i = 1 \mid x_i, \theta_i)}{\mathbb{P}(Y_i = 0 \mid x_i, \theta_i)} &=\log p(x_i \mid \theta_i) - \log p(x_i) + \log \frac{\mathbb{P}(Y_i=1)}{\mathbb{P}(Y_i=0)}
\end{align*}

Here, $p(x_i, \theta_i) = \sum_{c=0}^1 p(x_i, \theta_i \mid Y_i = c)\mathbb{P}(Y_i=c)$ is just the marginalization over $y_i$ per the law of total probability. The unconditional probability $\mathbb{P}(Y_i=1)$ is known and conventionally set to 0.5. Thus, the output of $h_\phi(x, \theta)$ gives an estimate of the log LER.

\section{DETAILED TRAINING PROCEDURE USING \ours}
\label{app:training}

Here, we present our training algorithm using score augmentation, adaptive weighting, and finite differencing. Unless otherwise specified, all floating point quantities are calculated using 64-bit precision.

First, a simulated dataset $\mathcal{D} = \{(x_i, \theta_i)\}_{i=1}^N$ is constructed for a given $N$ using Latin hypercube sampling of the parameter space. To do so, we partition the $d$-dimensional $\theta$ space into equally sized cells, with $\lceil N^{1/d} \rceil$ cells in each dimension. We then sample one $\theta$ value from each cell, and sample one $x\sim p(x\mid\theta)$ per $\theta$. This forms the $y=1$ class for our binary classification approach. We use this Latin hypercube sampling to ensure coverage of the parameter space and reduce variance between training runs. The training then proceeds according to Algorithm~\ref{alg:training}.

\begin{algorithm}[H]
\caption{Efficient Neural Likelihood Training using Score-Augmented Loss}\label{alg:training}
\KwIn{Simulated dataset $\mathcal{D} = \{(x_i, \theta_i)\}_{i=1}^N$ where $x_i \sim p(x \mid \theta_i)$ \\ \hspace{10mm} History weighting function $f_H$ (e.g., recency-weighted average) \\ \hspace{10mm} $\alpha$ update interval $T_\alpha$}
\KwOut{Trained surrogate model $h_\phi(x, \theta)$}
\BlankLine
 Initialize classifier $h_\phi(x, \theta)$ (outputs unnormalized logits)\;
 Initialize $\alpha \gets 0,\: t_\alpha \gets 0$\;
 Initialize history $H \gets \emptyset$\;
 Select finite difference perturbations $\epsilon_k \in \mathbb{R}^d$ to match $\nabla_\theta h_\phi(x, \theta)$\;

 \While{not converged}{
     \For{batch indices $B \subset [N]$}{
        $y = 1$ class: $B^1 \gets \{(x_i, \theta_i, 1) : i \in B\}$\;
        $y = 0$ class: $B^0 \gets \{(x_i, \theta_{\pi(i)}, 0) : i\in B\}$ where $\pi(i) \sim $ Uniform$\{1,2, ..., N\}$\;
        \For{$i \in B$}{
            \For{$k\in [d]$}{
                estimated score $s^k_i = [h_\phi(x_i, \theta_i + \epsilon_k \: e_k) - h_\phi(x_i, \theta_i)]/\epsilon_k$\;
            }
        }
        BCE loss $L_{BCE} \gets \sum_{(x_j,\theta_j, Y_j)\in B^1\cup B^0}\text{BCE}(h_\phi(x_j, \theta_j), Y_j)$\;
        Score loss $L_{Score} \gets \sum_{i\in B}\sum_{k=1}^d\alpha[s^k_i-\nabla_{\theta^k}\log p(x_i \mid \theta_i)]^2$\;
        Update $\phi$ using $\nabla_\phi(L_{BCE} + L_{Score})$\;
        $t_\alpha \gets t_\alpha + 1$\;

        \BlankLine
        \If{$t_\alpha \geq T_\alpha$}{
            $g^{BCE} \gets \nabla_\phi L_{BCE}$\;
            $g^{Score} \gets \nabla_\phi L_{Score}$\;
            $\alpha' \gets \frac{\|g^{BCE}\|_2}{\|g^{Score}\|_2}$\;
            $H\gets H \cup \{\alpha'\}$\;
            $t_\alpha \gets 0$\;
        }

        $\alpha \gets f_H(H)$\;

     }
}
 \Return{$h_\phi(x, \theta)$}\;
\end{algorithm}

For brevity, we have omitted the usual hyperparameters for training neural networks, such as batch size, learning rate, and optimizer configurations (see Appendix~\ref{app:exp_spec} for more information).

\subsubsection{Automatic Tuning of Finite Difference Predicted Score}

For finite difference perturbations $\epsilon_k$, we first calculate the ground truth predicted score $\nabla^*_\theta h_\phi(x, \theta)$ over the first minibatch $B$ of training data using standard backpropagation. Starting with $\epsilon_k$ values around $10^{-5}$, we calculate the maximum relative error

\begin{align*}
    \max_{i\in B, k} \left|\frac{\nabla^*_{\theta^k} h_\phi(x_i, \theta_i) - \nabla^{\epsilon_k}_{\theta^k} h_\phi(x_i, \theta_i)}{\nabla^*_{\theta^k} h_\phi(x_i, \theta_i)}\right|
\end{align*}

where $\nabla^{\epsilon_k}_{\theta^k} h_\phi(x, \theta)$ is the finite difference approximation to $\nabla^*_{\theta^k} h_\phi(x, \theta)$ using $\epsilon_k$. If the maximum relative error exceeds 0.01, we reduce the $\epsilon_k$ values and recalculate on the same minibatch until the maximum relative error is below the threshold. If the threshold is never reached, we try again on the next minibatch of training data until we find a suitable set of $\epsilon_k$ values.

\subsection{Details on sparse $\alpha$ update procedure}

In all experiments, we set $T_\alpha = 64$ and define $f_H$ below. Let $\alpha_{t_i}$ be the $\alpha$ value calculated at cumulative batch $t_i$. Then, at the current batch $t_0$, we have

\begin{align*}
    f_H(H) = \frac{\sum_{\alpha_{t_i} \in H(64)} \alpha_{t_i} \exp\left(-\frac{t_0 - t_i}{64}\right)}{\sum_{\alpha_{t_i} \in H(64)} \exp\left(-\frac{t_0 - t_i}{64}\right)}
\end{align*}

where $H(64)$ are the most recent 64 $\alpha$ values in $H$. In essence, we use an exponential recency-weighted moving average of historical $\alpha$ values in the calculation for \ours.

\section{HARDWARE AND SOFTWARE SPECIFICATIONS}
\label{app:hardware}
All neural network training was performed on NVIDIA L40 GPUs (48~GB of VRAM) and AMD EPYC 7763 64-core processors. Our simulation and neural network optimization code relies heavily on PyTorch \citep{10.5555/3454287.3455008}, available under the BSD 3-Clause License.

\section{EXPERIMENTAL SPECIFICATIONS}
\label{app:exp_spec}

This section provides additional details for specifications and training hyperparameters used in experiments.

\subsection{Definition of performance metrics}
\label{app:perf_metrics_def}

Evaluation of each surrogate likelihood is done in two stages. First, we construct the ``L-test'' set by sampling 100,000 $(x_i, \theta_i)$ tuples, where $\theta_i$ is sampled uniformly from a fixed parameter space specific to each experimental setting (see Appendix \ref{app:exp_spec}) and $x_i \sim p(x \mid \theta_i)$. This initial test set and an additional 100,000 tuples where $x_i$ and $\theta_i$ are permuted (see Appendix~\ref{app:classifier_trick}) are used calculate $L_{BCE}$. We then use the initial 100,000 paired samples to additional evaluate $L_{Score}$. Both metrics measure how closely the NRE-A surrogate approximates the true LER from an optimization perspective. Across all settings in a case study, the L-test set is identical.

Second, we generate the ``E-test'' set by constructing an evenly spaced, roughly 100-point grid $\{\theta_i\}$ in the parameter space. For each $\theta_i$, we sample 30 groups of 50 data points $\mathbf{x_{i, g}} = \{x_{i, g, j}\}$, where $x_{i, g, j} \overset{\text{iid}}{\sim} p(x \mid \theta_i)$. This test set is used to evaluate performance of the surrogate on the following downstream inference tasks:

\begin{itemize}
    \item \textbf{Hypothesis testing:} We compute the null likelihood ratio test statistic $\Lambda(\mathbf{x_{i, g}}, \theta_i) = 2\log p(\mathbf{x_{i, g}}, \hat \theta) - 2\log p(\mathbf{x_{i, g}}, \theta_i)$, where $\hat \theta$ is the maximum likelihood estimate (MLE) of $\theta$ given $\mathbf x_{i, g}$. We examine the aggregate and distributional discrepancy of surrogate $\Lambda$ values vs. ground truth values.
    \item \textbf{Confidence sets:} We generate 95\% Wilks' theorem-based confidence sets for $\theta$ given $\mathbf{x_{i, g}}$ and compute their empirical coverage and average size. Specifically, for each $\mathbf{x_{i, g}}$ we form the set $\hat{C}(\mathbf{x}_{i,g}) = \{\theta : 2[h_\phi(\hat{\theta}; \mathbf{x}_{i,g}) - h_\phi(\theta; \mathbf{x}_{i,g})] \le \chi^2_{d,\,0.95}\}$, where $\hat{\theta}$ is the surrogate MLE, and $\chi^2_{d,\,0.95}$ is the $0.95$ quantile of a chi-squared distribution with $d = \dim(\theta)$ degrees of freedom. By Wilks' theorem, the likelihood ratio statistic evaluated at the true parameter is asymptotically $\chi^2_d$, so this set attains approximately nominal $95\%$ coverage when the likelihood is well approximated.
    \item \textbf{Point estimation:} We compute the MLE $\hat \theta = \arg\max_\theta p(\mathbf{x_{i, g}} \mid \theta)$ and measure average discrepancy between surrogate MLEs and ground truth MLEs.
\end{itemize}

Although the surrogate $h_\phi(x, \theta)$ only takes in one $x$ value per $\theta$ value (sample size 1), we use the additivity of log likelihoods to construct LR test statistics, confidence sets, and MLEs for sample size 50 by directly summing surrogate log likelihoods. This is done to achieve sufficient sample size for LRTS and Wilks' confidence set asymptotics. This also has the nice side effect of accentuating any mis-calibration of the surrogate likelihood. Small errors at sample size 1 compound and become more visible at sample size 50, making comparisons between surrogates easier.

\subsection{Common experimental specifications}

Below are details of experiment settings common to each case study.

\begin{itemize}
    \item We vary dataset size $N\in \{30\text{k}, 100\text{k}, 300\text{k}, 1\text{m}\}$ for all experimental settings (k = 1,000, m = 1,000,000). See Appendix \ref{app:training} for how the simulated datasets are generated.
    \item A fixed, bounded base parameter space for $\theta$ values is used for both surrogate training and evaluation. For training data, we slightly extend the base parameter space. For E-test set generation, we add an inner margin to the base parameter space (this does not apply to the L-test set). This is done to avoid boundary effects. See Appendices \ref{app:sis_spec}, \ref{app:gn_spec}, and \ref{app:stp_spec} for exact parameter space bounds.
    \item The sequence of layer types in the neural network is fixed, but the size of each network is varied by adjusting layer widths. See Appendices \ref{app:sis_spec}, \ref{app:gn_spec}, and \ref{app:stp_spec} for exact neural network architectures.
    \item We use the AdamW~\citep{loshchilov2019decoupledweightdecayregularization} optimizer with a fixed learning rate and batch size 64 for each case study. We train the neural network for a minimum number of epochs then stop using convergence patience of 5 epochs on validation BCE loss. The minimum epoch requirement is used to prevent convergence patience from ending a training run too early. The validation dataset and training dataset are of equal size and generated using the same procedure in Appendix \ref{app:training} under different random seeds. See Appendices \ref{app:sis_spec}, \ref{app:gn_spec}, and \ref{app:stp_spec} for exact values of learning rate and see Table~\ref{tab:min_epochs} for minimum epochs corresponding to each dataset size.

\end{itemize}

\begin{table}[h!]
  \centering
  \caption{NLER training hyperparameters}
  \label{tab:min_epochs}
  \begin{tabular}{|l|r|}
    \hline
    \textbf{Dataset Size} & \textbf{Minimum Training Epochs} \\
    \hline
    30K & 40  \\
    100K & 30 \\
    300K & 20 \\
    1M & 10 \\
    \hline
  \end{tabular}
\end{table}

\subsection{Specifications and hyperparameters for continuous-time hidden Markov model for SIS Epidemics}
\label{app:sis_spec}

We model a small population using $K=8$ nodes positioned on the 2D Euclidean plane. Four nodes are positioned at the vertices of the unit square $[(0, 0), (1, 0), (0, 1), (1, 1)]$, and the other four are positioned at a copy of the unit square translated +2 units in both the x and y directions at $[(2, 2), (3, 2), (2, 3), (3, 3)]$. We construct a fully connected graph from these nodes and assign edge weights $w_{i,j} = \exp(-d_{i,j})$, where $d_{i,j}$ is the Euclidean distance between nodes $i$ and $j$.

Let $v_{k, t} \in \{0, 1\}$ indicate the state of node $k$ at time $t$, either susceptible (0) or infected (1). We use a continuous-time Markov chain (CTMC) to model intrinsic transitions in our graph. Given node state $v_{k, t}$, let $\Delta t_{k}$ be time remaining until the next intrinsic transition for node $k$. The CTMC is then defined via rate-parameterized exponential distributions:
\begin{align*}
    \Delta t_{k} \mid (v_{k, t}=0) &\sim \text{Exp}\left(\eta + \lambda \sum_{j=1}^K w_{k,j} v_{j, t} \right) \\ 
    \Delta t_{k} \mid (v_{k, t}=1) &\sim \text{Exp}\left(\mu\right),
\end{align*}
where the (external) infection rate $\lambda$, spontaneous infection rate $\eta$, and recovery rate $\mu$ are the SPM parameters of interest. We fix  $\eta=0.135$ and perform inference on $\lambda$ and $\mu$ only. 

Next, we assume that we do not observe the graph at every time $t$, but at some fixed, finite subset of times $\{t_i\}_{i=1}^T$. Let $x^\text{obs}_{t_i}$ be the observed state of the graph at time $t_i$, represented as a single integer in $\{1, ..., 2^K\}$. Under this continuous-time hidden Markov model (CT-HMM), intrinsic graph transitions happen one node at a time, but $x^\text{obs}_{t_i}$ and $x^\text{obs}_{t_{i+1}}$ may differ by any number of nodes. Furthermore, multiple trajectories of intrinsic graph transitions could have occurred in between measurements. See Figure~\ref{fig:data} for a visual of how intrinsic transition times interact with measurement times.

Per Equation~\eqref{eq:sis_likelihood}, the likelihood function for the SIS epidemic model is given by

\begin{align*}
p( \{x_{t_i}^\text{obs}\}_{i=1}^T\mid\lambda, \mu, \eta ) &= P(x^\text{obs}_{t_1}) \prod_{i=2}^T \left[ \exp \left( Q(\lambda, \mu, \eta) (t_i - t_{i-1}) \right) \right]_{x^\text{obs}_{t_{i-1}}, x^\text{obs}_{t_i}}
\end{align*}

where $\lambda$ is the infection rate, $\mu$ is the recovery rate, $\eta$ is the self-infection rate, and $\{x_{t_i}^\text{obs}\}_{i=1}^T$ is an observed sequence of graph states. The generator matrix $Q$ encodes the instantaneous rates of changing from state $x^\text{obs}_{t_{i-1}}$ to state $x^\text{obs}_{t_i}$ and is defined as
-
\begin{align*}
    Q_{x^\text{obs}_{t_{i-1}}, x^\text{obs}_{t_i}} (\lambda, \mu, \eta) &= \begin{cases}
\mu & x^\text{obs}_{t_{i-1}} \text{ becomes } x^\text{obs}_{t_i} \text{ via  node $k$ changing } 1 \to 0 \\
\eta + \lambda \sum_{j=1}^K w_{k,j} v_{j, t_{i-1}} & x^\text{obs}_{t_{i-1}} \text{ becomes } x^\text{obs}_{t_i} \text{ via node $k$ changing } 0 \to 1 \\
0 & x^\text{obs}_{t_{i-1}} \text{ and } x^\text{obs}_{t_i} \text{ differ by more than one node} \\
-\sum_{y \neq x^\text{obs}_{t_i}} Q_{x^\text{obs}_{t_i}, y} &  x^\text{obs}_{t_{i-1}} = x^\text{obs}_{t_i}
\end{cases}
\end{align*}

The matrix exponential in the likelihood definition accounts for all possible paths the graph could have taken between $x^\text{obs}_{t_{i-1}}$ and $x^\text{obs}_{t_i}$. The parameter space used in the SIS epidemics case study is given in Table~\ref{tab:sis_params}.

\begin{table}[h!]
  \centering
  \caption{Parameter space for SIS epidemics case study}
  \label{tab:sis_params}
  \begin{tabular}{|l|l|l|r|r|}
    \hline
    \textbf{Raw Parameter} & \textbf{Working Parameter} & \textbf{Bounds} & \textbf{Train Margin} & \textbf{E-test Margin} \\
    \hline
    Infection Rate ($\lambda$)  & $\log \lambda$ & [-1, 1] & 20\% & 40\% \\
    Recovery Rate ($\mu$)   & $\log \mu$ & [-1, 1] & 20\% & 40\% \\
    Self-infection rate ($\eta$) & $\eta$ & = 0.135 & N/A & N/A  \\
    \hline
  \end{tabular}
\end{table}

The bounds in Table~\ref{tab:sis_params} apply to the working parameter, e.g. $\log \lambda \in [-1, 1]$ defines the base parameter space for $\lambda$. The train and E-test margins are split equally; for example, the effective parameter space used in NLER training for $\lambda$ is $\log \lambda \in [-1.1, 1.1]$. In addition to these SIS parameters, we set measurement times $\{t_i\}_{i=1}^T = \{0, 1, \dots, 12\}$.

The neural network $h_\phi(x, \theta)$ takes input shape [batch size, $T$, $K$], where the last dimension is the one-hot encoding of the graph state (a value of 1 in the $k$th position indicates that node $k$ is currently infected). It is passed through a shallow MLP to learn a state embedding, yielding a shape [batch size, $T$, embedding dimension]. Then, the last two dimensions are transposed and the input is fed through a sequence of three 1D convolution layers (kernel size 5) to yield a final shape of [batch size, feature size]. The features are concatenated with raw parameter values to yield [batch size, feature size + $\theta$ dimension], then the resulting tensor is fed through a series of linear layers to yield logits of shape [batch size, 1]. All layers except the final linear layer are followed with SiLU (sigmoid linear unit) activations. Table~\ref{tab:sis_arch} shows the exact architectures for each surrogate size.

\begin{table}[h!]
  \centering
  \caption{Surrogate architectures for SIS epidemics case study}
  \label{tab:sis_arch}
  \begin{tabular}{|l|r|r|r|r|}
    \hline
    \textbf{Layer Type} & \textbf{Size 3K} & \textbf{Size 10K} & \textbf{Size 30K} & \textbf{Size 100K} \\
    \hline
    Embedder hidden units  & [16, 8] & [32, 16] & [64, 32] & [64, 64]\\
    Convolution output channels   & [8, 12, 16] & [16, 24, 24] & [32, 32, 48] & [64, 64, 96] \\
    Linear layers hidden units & [30, 12] & [52, 32] & [64, 64, 32] & [128, 64, 32]  \\
    \hline
  \end{tabular}
\end{table}

All surrogates in this case study are trained using learning rate $10^{-3}$ with weight decay $10^{-6}$.

\subsection{Specifications and hyperparameters for anisotropic gaussian spatial process with nugget effect}
\label{app:gn_spec}
Let $\mathbf{Z}$ be a 2D Gaussian process (GP) with an anisotropic covariance kernel parameterized by length scales $l_x$ and $l_y$, one for each spatial dimension. Given two sampling locations $s, s' \in \mathbb{R}^2$, the spatial covariance of the random GP field $\mathbf{Z}$ at those points is given by

\begin{align}
    \text{Cov}(\mathbf{Z}(s), \mathbf{Z}(s') \mid l_x, l_y) &= \sigma^2 \exp \left( - \sqrt{ \frac{(s_1 - s'_1)^2}{l_x^2} + \frac{(s_2 - s'_2)^2}{l_y^2} } \right)
\end{align}

where $l_x$ and $l_y$ are the length scale parameters for each spatial direction, and $\sigma^2$ is the GP variance (fixed at $\sigma^2 = 1$). Larger/smaller length scales will define fields that vary more slowly/rapidly over space in the given direction. See Figure~\ref{fig:data} for examples of field samples for different length scales.

We also add a small nugget variance parameter $\varepsilon^2$ to model small-scale variations that are independent across spatial locations. A GP observed at $S$ sample locations $\mathbf{s} =\{s_i\}_{i=1}^S$ with this added nugget variance can be parameterized by a covariance matrix

\begin{align*}
    \Sigma(l_x, l_y, \varepsilon)_{i, j} &= \text{Cov}(\mathbf{Z}(s_i), \mathbf{Z}(s_j)\mid l_x, l_y) + \varepsilon^2 \mathbf{1}(i=j)
\end{align*}

The likelihood given $\mathbf{x}^\text{obs}$ is then defined  as in Equation~\eqref{eq:gp_likelihood}
\begin{align*}
p(\vxobs \mid l_x, l_y, \varepsilon) &= (2\pi)^{-S/2} |\Sigma(l_x, l_y, \varepsilon)|^{-1/2}\exp\left( -\frac{1}{2} (\vxobs)^\top \Sigma(l_x, l_y, \varepsilon)^{-1} \vxobs \right).
\end{align*}

The bounds of the parameter space used in the GP case study are given in Table~\ref{tab:gn_params}.

\begin{table}[h!]
  \centering
  \caption{Parameter space for Gaussian process case study}
  \label{tab:gn_params}
  \begin{tabular}{|l|l|l|r|r|}
    \hline
    \textbf{Raw Parameter} & \textbf{Working Parameter} & \textbf{Bounds} & \textbf{Train Margin} & \textbf{E-test Margin} \\
    \hline
    GP variance ($\sigma^2$)  & $\sigma^2$ &  = 1 & N/A & N/A \\
    $x$-length scale ($l_x$)   & $\log l_x$ & [-1, 1] & 10\% & 40\% \\
    $y$-length scale ($l_y$)   & $\log l_y$ & [-1, 1] & 10\% & 40\% \\
    Nugget variance ($\varepsilon^2$) & $\log \varepsilon$ & [-4, -1] & 10\% & 40\%  \\
    \hline
  \end{tabular}
\end{table}

The bounds in Table~\ref{tab:gn_params} apply to the working parameter, e.g. $\log l_x \in [-1, 1]$ defines the base parameter space for $l_x$. The train and E-test margins are split equally on each side of the parameter space; for example, the effective parameter space used in surrogate training for $l_x$ is $\log l_x \in [-1.05, 1.05]$. In addition to these GP parameters, we define our sample points as a $25\times25$ uniform grid across the subset of the 2D plane $[-3, 3] \times [-3, 3]$.

The surrogate network takes an input of shape [batch size, 1, 25, 25], representing the spatial field as an image. It is passed through a sequence of three 2D convolution layers (kernel size 3), each followed by an average pooling layer (kernel size 2, stride 2) to yield a final shape of [batch size, feature size]. The features are concatenated with raw parameter values to yield [batch size, feature size + $\theta$ dimension], then the resulting tensor is fed through a series of linear layers to yield logits of shape [batch size, 1]. All convolutional layers are followed by ReLU (rectified linear unit) activations, while all linear layers except the last are followed by SiLU (sigmoid linear unit) activations. Table~\ref{tab:gp_arch} shows the exact architectures for each surrogate size.

\begin{table}[h!]
  \centering
  \caption{Surrogate architectures for Gaussian process case study}
  \label{tab:gp_arch}
  \begin{tabular}{|l|r|r|r|r|}
    \hline
    \textbf{Layer Type} & \textbf{Size 30K} & \textbf{Size 100K} & \textbf{Size 300K} & \textbf{Size 1M} \\
    \hline
    Convolution output channels   & [40, 40, 32] & [80, 80, 44] & [128, 128, 108] & [256, 256, 136] \\
    Linear layers hidden units & [48, 32] & [64, 64, 32] & [128, 80, 64] & [256, 176, 96]  \\
    \hline
  \end{tabular}
\end{table}

All surrogates are trained using learning rate $10^{-3}$ with weight decay $10^{-6}$.

\subsection{Specifications and hyperparameters for anisotropic Student-$t$ spatial process}
\label{app:stp_spec}

A Student-$t$ spatial process (STP) is parameterized by a covariance matrix similarly to a Gaussian process (GP). In addition, the sampling process for an STP is analogous to the univariate case. One can first sample an ordinary GP, then scale the entire field by a $\chi_\nu^2$ random variate. See Equation~\eqref{eq:stp_app} below for the sampling process and the resulting likelihood.

\begin{align}
\begin{split}
\label{eq:stp_app}
    \mathbf{Z} &\sim \text{GP}(0, \Sigma(l_x, l_y, \varepsilon)) \\
    R &\sim \chi^2_\nu \text{ (univariate sample)} \\
    \mathbf{X} &= \mathbf{Z}\sqrt{\frac{\nu-2}{R}} \\
    p(\vxobs \mid \theta) &= \frac{\Gamma(\frac{\nu + S}{2})}{\pi^{S/2}(\nu-2)^{S/2}\Gamma(\nu/2)}|\Sigma(l_x, l_y, \varepsilon)|^{-1/2}\left(1+\frac{(\vxobs)^\top\Sigma(l_x, l_y,\varepsilon)^{-1}\mathbf{x}^\text{obs}}{\nu-2}\right)^{-\frac{\nu+S}{2}}
\end{split}
\end{align}

The conventional STP sampling process scales the GP by $\sqrt{\nu/R}$. Our $\sqrt{(\nu-2)/R}$ scaling ensures that the covariance of $\mathbf{X}$ is $\Sigma(l_x, l_y, \varepsilon)$ instead of $\frac{\nu}{\nu-2}\Sigma(l_x, l_y, \varepsilon)$, which eliminates the dependence of the overall scale of $\mathbf{X}$ on $\nu$ but restricts our parameter space to $\nu>2$.

The bounds of the parameter space used in the STP case study are given in Table~\ref{tab:stp_params}.

\begin{table}[h!]
  \centering
  \caption{Parameter space for Student-$t$ process case study}
  \label{tab:stp_params}
  \begin{tabular}{|l|l|l|r|r|}
    \hline
    \textbf{Raw Parameter} & \textbf{Working Parameter} & \textbf{Bounds} & \textbf{Train Margin} & \textbf{E-test Margin} \\
    \hline
    STP variance ($\sigma^2$)  & $\sigma^2$ &  = 1 & N/A & N/A \\
    $x$-length scale ($l_x$)   & $\log l_x$ & [-1, 1] & 10\% & 40\% \\
    $y$-length scale ($l_y$) & $\log l_y$ & [-1, 1] & 10\% & 40\% \\
    Nugget variance ($\varepsilon^2$) & $\varepsilon$ & = 0 & N/A & N/A \\
    Degrees of freedom ($\nu$) & $\log (\nu - 2)$ & [-2, 3] & 10\% & 40\% \\
    \hline
  \end{tabular}
\end{table}

The bounds in Table~\ref{tab:stp_params} apply to the working parameter, e.g. $\log l_x \in [-1, 1]$ defines the base parameter space for $l_x$. The train and E-test margins are split equally; for example, the effective parameter space used in surrogate training for $l_x$ is $\log l_x \in [-1.05, 1.05]$. In addition to these STP parameters, we define our sample points as a 25x25 uniform grid across the subset of the 2D plane $[-3, 3] \times [-3, 3]$.

The surrogate architecture for the STP case study is identical to that used for the GP case study (see Table~\ref{tab:gp_arch}). All surrogates are trained using learning rate $3\times10^{-3}$ with weight decay $10^{-6}$.

\section{ADDITIONAL RESULTS}
\label{app:additional_results}
\subsection{Case study 1: Continuous-time hidden Markov model for SIS Epidemics (SIS CT-HMM)}
\label{app:sis_results}

\begin{figure}[ht]
    \centering
    \includegraphics[width=0.6\linewidth]{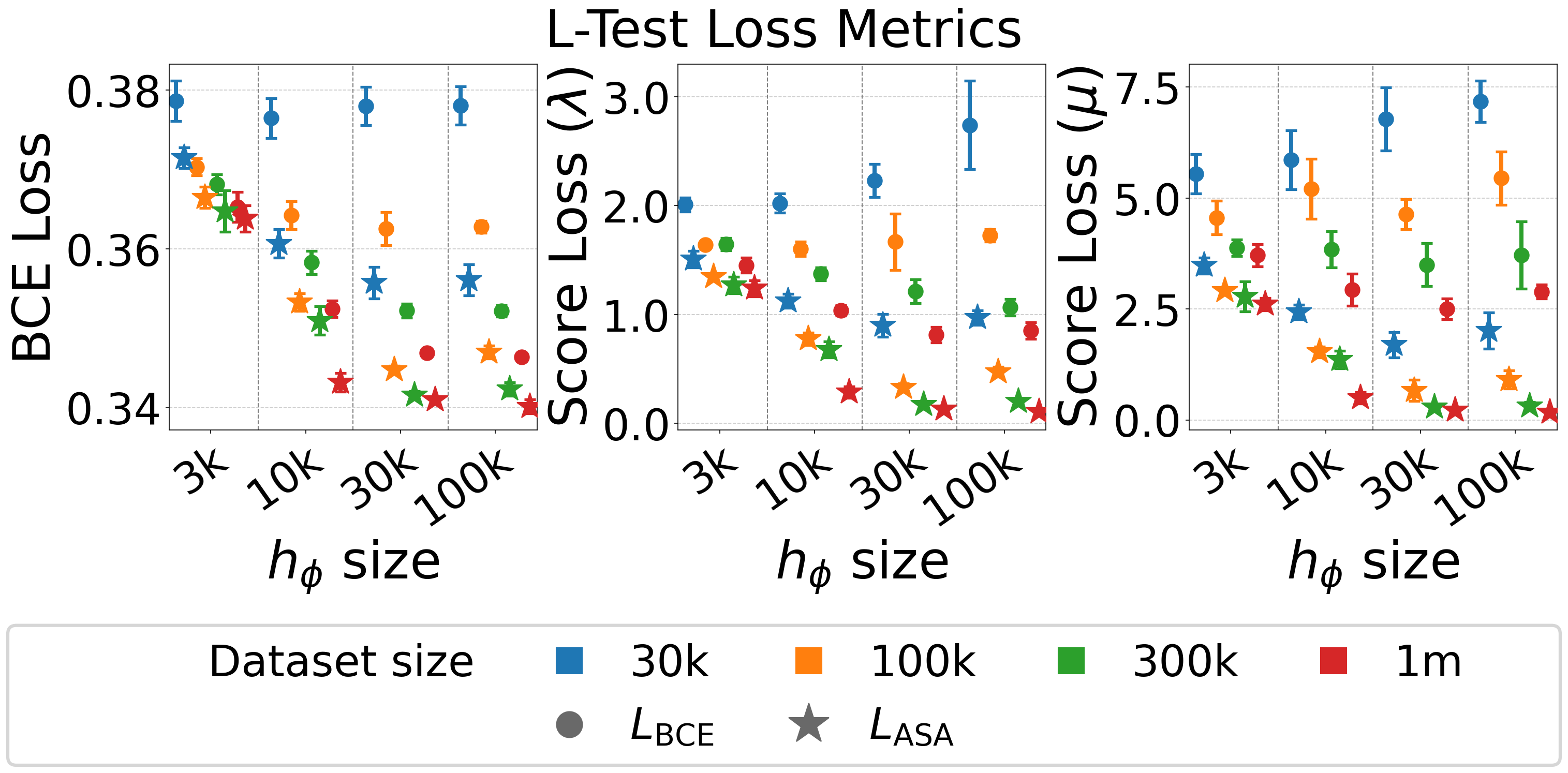}
    \caption{\textbf{CT-HMM for SIS epidemics test BCE and score loss}: For each $h_\phi(x, \theta)$ size (vertical lanes) and training dataset size (marker colors), we compute test BCE \textbf{(Left)}, test score loss for infection rate \textbf{(Middle)}, and test score loss for recovery rate \textbf{(Right)} for the surrogate trained under $L_{BCE}$ (dot) vs.\ trained under $L_{ASA}$ (star). Quantities are means $\pm2$ standard errors across 5 seeds.}
    \label{fig:sis_1.5}
\end{figure}

\begin{figure}[ht]
    \centering
    \includegraphics[width=0.6\linewidth]{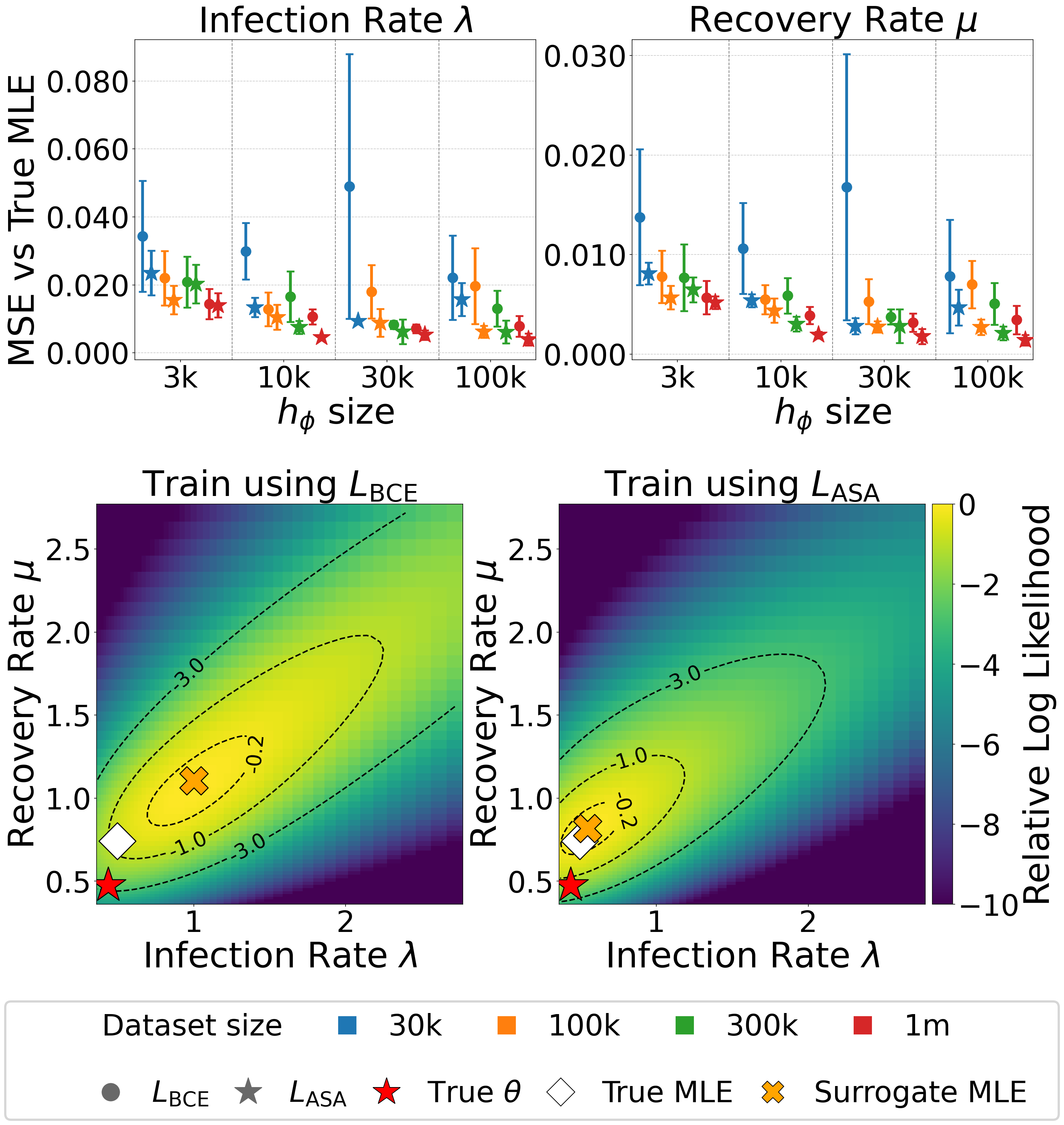}

    \caption{\textbf{CT-HMM for SIS epidemics maximum likelihood estimator (MLE) metrics and example likelihood surfaces: (Top)} MSE of surrogate MLE with respect to true MLE for same $\mathbf{x_{i,g}}$. Mean $\pm2$ standard errors over 5 seeds is plotted for each $h_\phi(x, \theta)$ size (vertical lanes), training dataset size (marker colors) and loss function (marker shape). \ours results in improved or unaffected MLE performance across most NRE-A settings. \textbf{(Bottom)} Example surrogate likelihood surfaces for same $x^\text{obs}$, scaled relative to maximum and with associated contours. Both surfaces use $h_\phi(x, \theta)$ size 100k and 100k training data points. We see that \ours modifies the shape of the entire surrogate likelihood surface, ultimately shifting the surrogate MLE closer to the ground truth MLE (white diamond).}
    \label{fig:sis_3}
\end{figure}

Figure~\ref{fig:sis_1.5} shows score loss metrics and Figure~\ref{fig:sis_3} shows MLE metrics and example likelihood surfaces for the SIS CT-HMM case study.

\subsection{Case study 2: Anisotropic Gaussian spatial process with nugget effect}
\label{app:gn_results}

\begin{figure}[ht]
    \centering
    \includegraphics[width=0.6\linewidth]{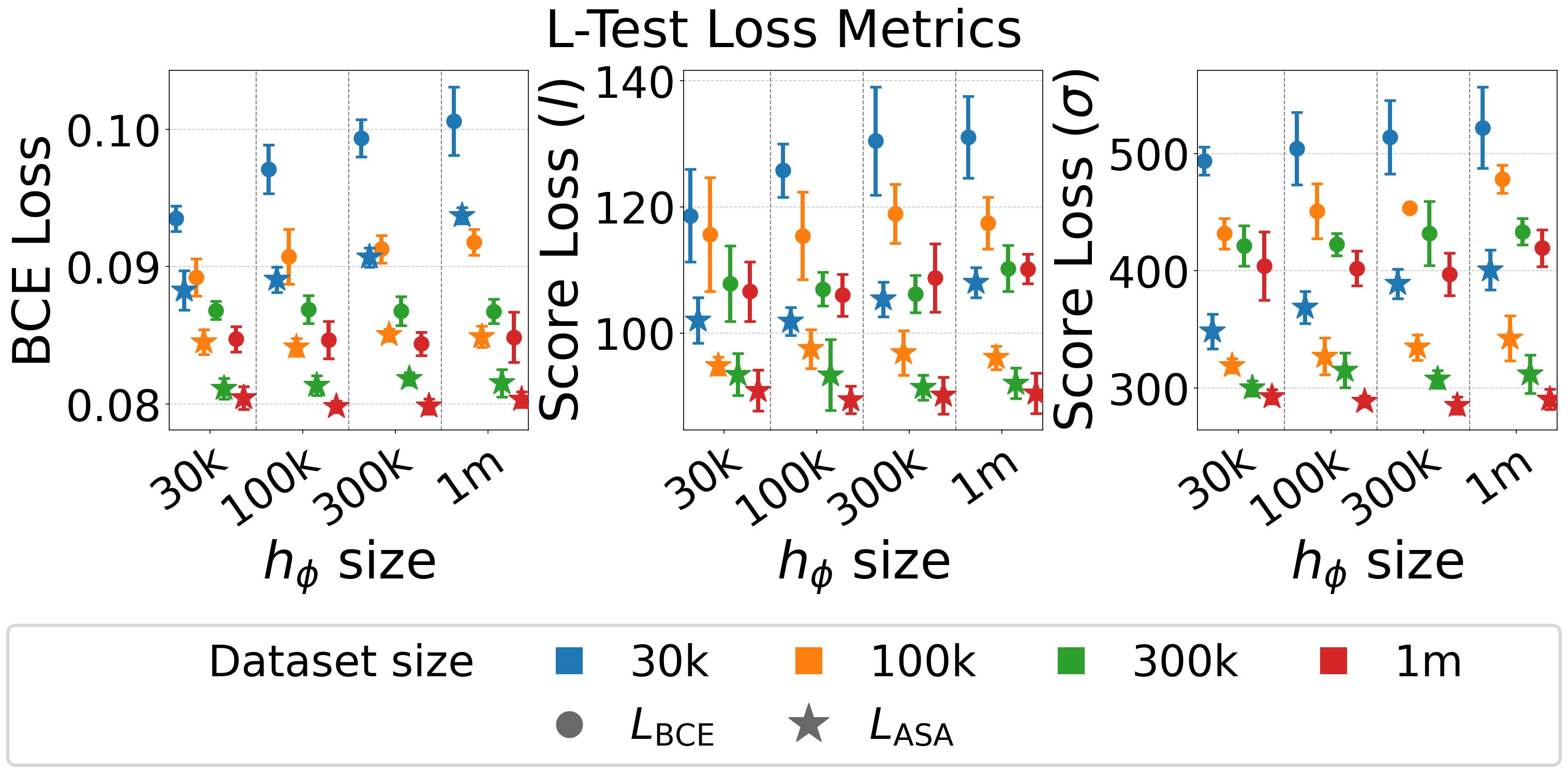}
    \caption{\textbf{Anisotropic Gaussian spatial process test BCE and score loss:} Figure is formatted identically to Figure~\ref{fig:sis_1.5}. Metrics for each length scale $(l_x, l_y)$ are combined in the middle figure}
    \label{fig:gn_1.5}
\end{figure}

Figure~\ref{fig:gn_1.5} shows score loss metrics for the Gaussian process case study. Figures \ref{fig:gn_2} and \ref{fig:gn_3} below show the E-test metrics for the Gaussian process case study.

\begin{figure}[ht]
    \centering
    \includegraphics[width=0.6\textwidth]{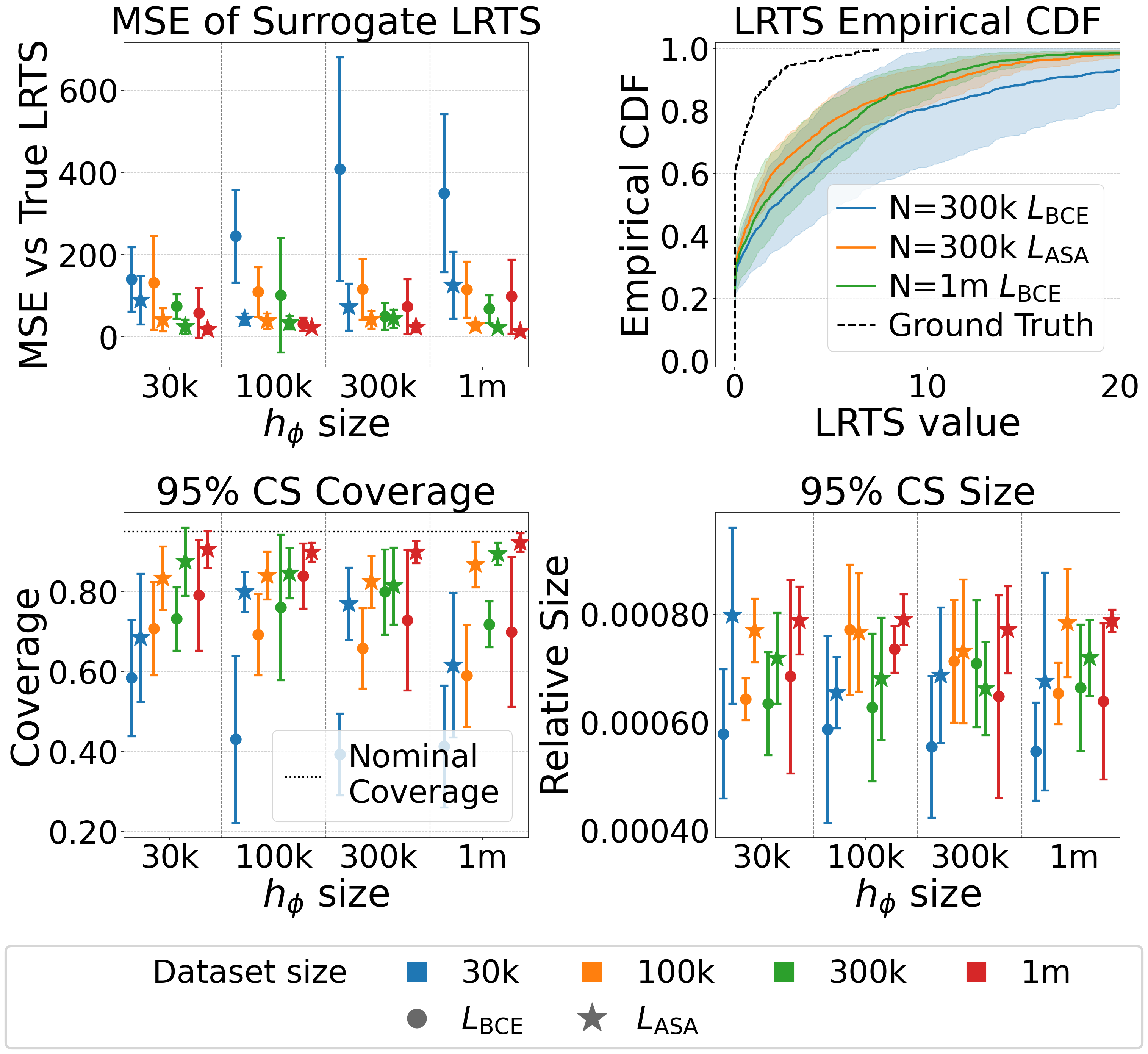}
    \caption{\textbf{Anisotropic Gaussian spatial process hypothesis testing and confidence set metrics for different NLER configurations:} Figure is formatted identically to Figure~\ref{fig:sis_2}}
    \label{fig:gn_2}
\end{figure}

\begin{figure}[ht]
    \centering
    \includegraphics[width=0.7\textwidth]{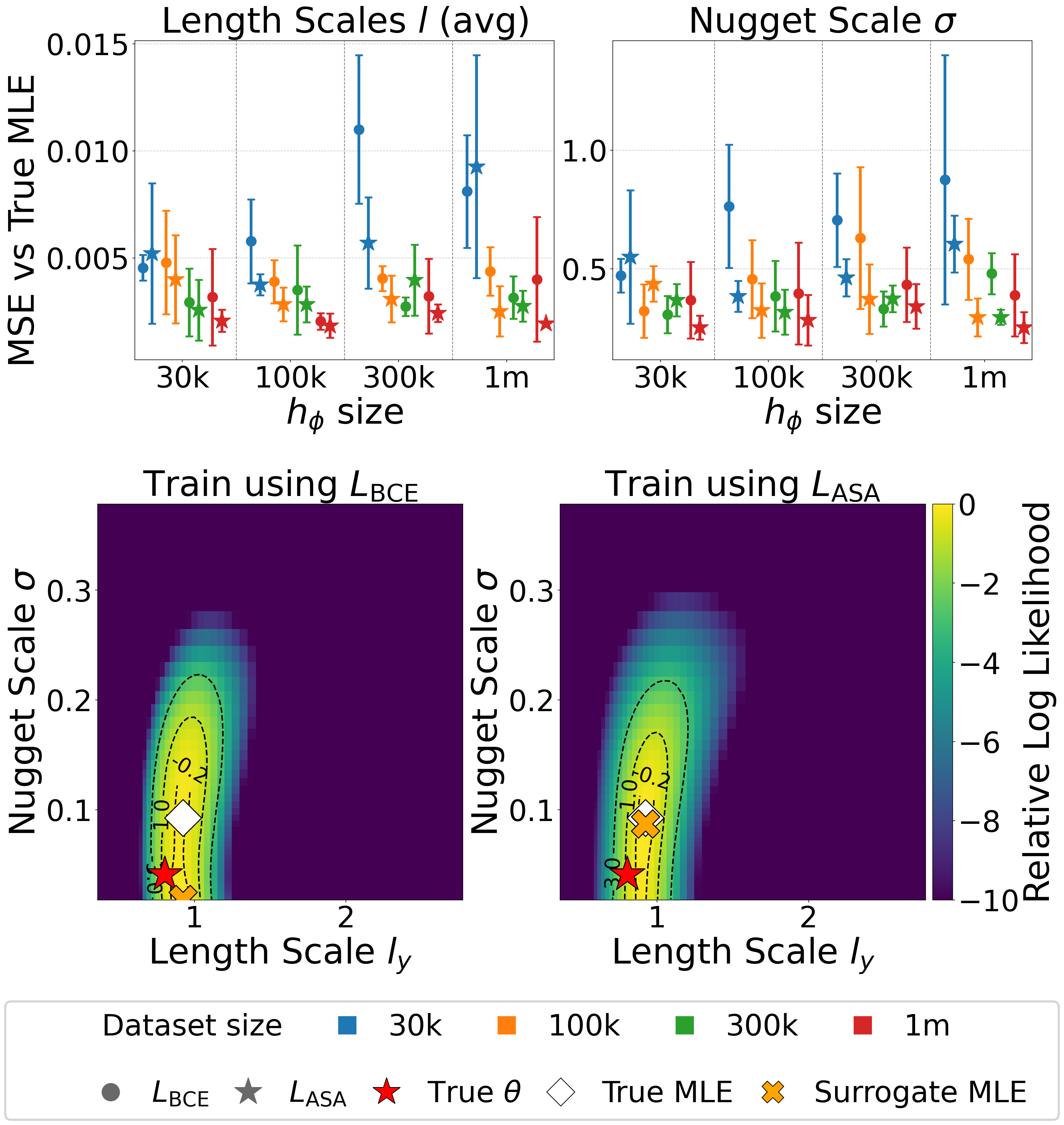}
    \caption{\textbf{Anisotropic Gaussian spatial process maximum likelihood estimator (MLE) metrics and example likelihood surfaces:} Figure is formatted identically to Figure~\ref{fig:sis_3}. MLE metrics for each length scale $(l_x, l_y)$ are averaged together. Likelihood surfaces and associated points are show for a slice at $l_x = 1$.}
    \label{fig:gn_3}
\end{figure}

\subsection{Case study 3: Anisotropic Student-$t$ spatial process}
\label{app:stp_results}

Figure~\ref{fig:stp_1.5} below shows score loss metrics, and Figures~\ref{fig:stp_2} and \ref{fig:stp_3} below show E-test metrics for the Student-$t$ spatial process case study.

\begin{figure}[ht]
    \centering
    \includegraphics[width=0.6\linewidth]{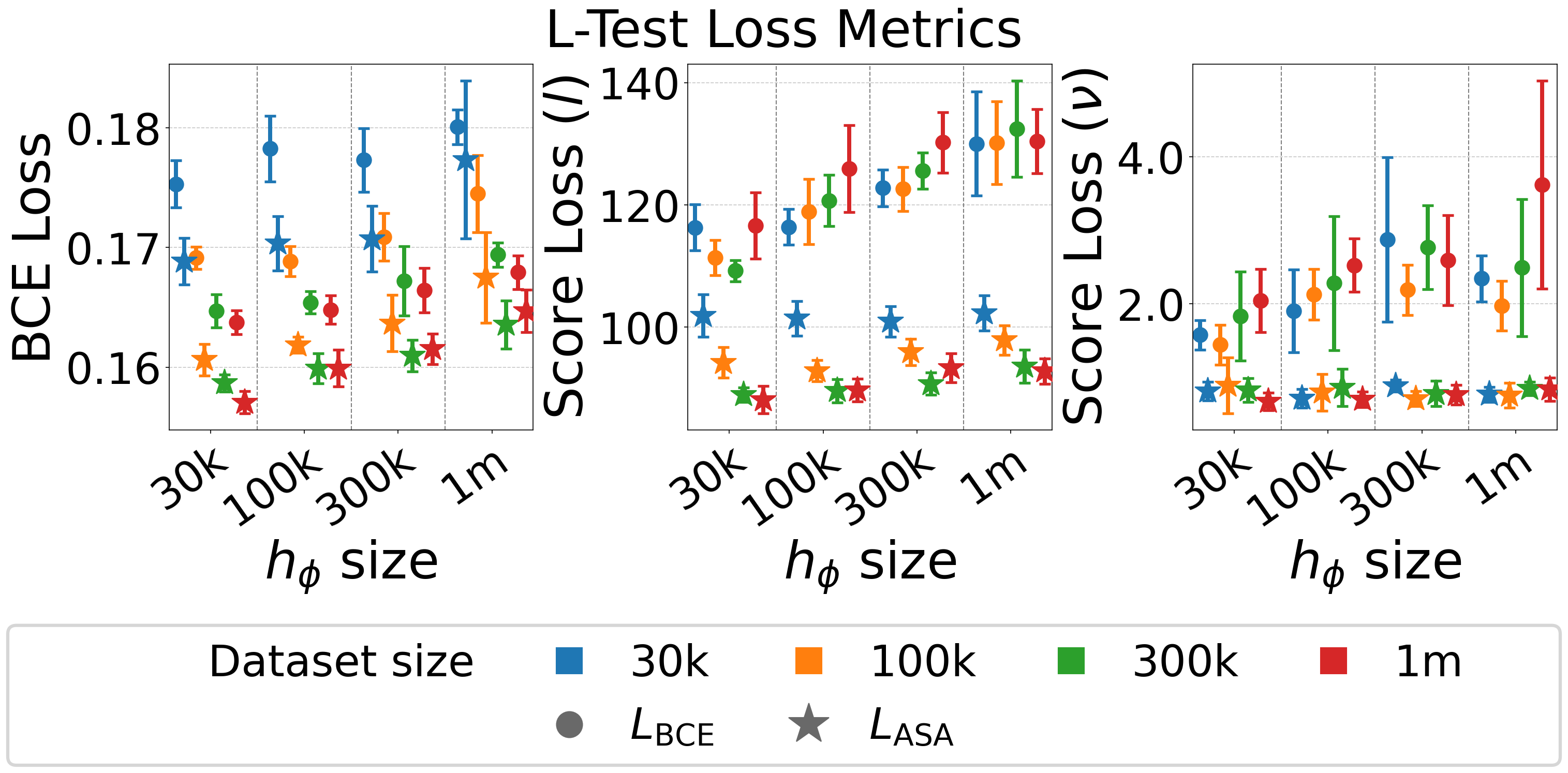}
    \caption{\textbf{Anisotropic Student-$t$ spatial process test BCE and score loss:} Figure is formatted identically to Figure~\ref{fig:sis_1.5}. Metrics for each length scale $(l_x, l_y)$ are combined in the middle figure}
    \label{fig:stp_1.5}
\end{figure}

\begin{figure}[ht]
    \centering
    \includegraphics[width=0.6\textwidth]{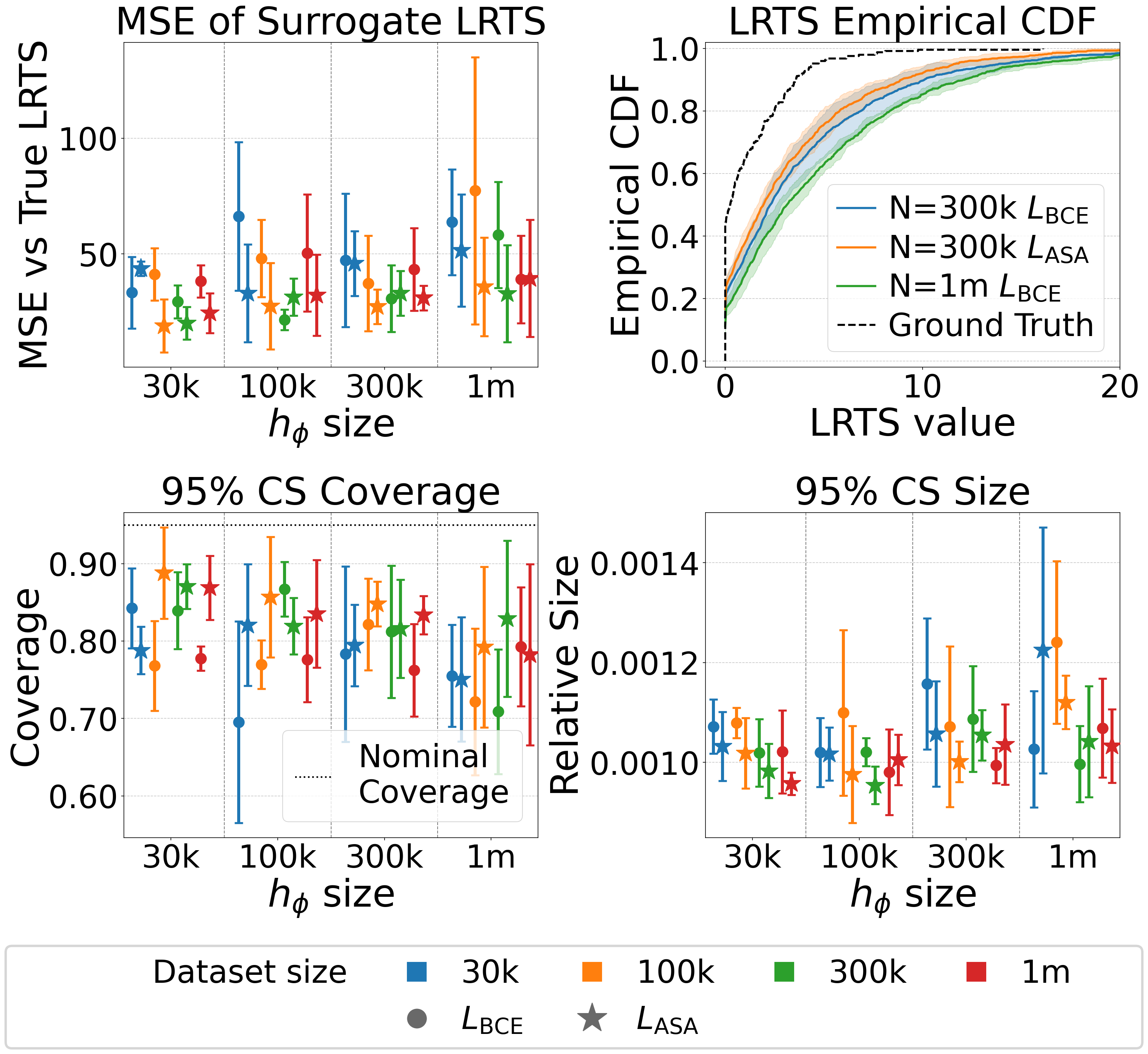}
    \caption{\textbf{Student-$t$ spatial process hypothesis testing and confidence set metrics for different surrogate settings:} Figure is formatted identically to Figure~\ref{fig:sis_2}.}
    \label{fig:stp_2}
\end{figure}

\begin{figure}[ht]
    \centering
    \includegraphics[width=0.6\textwidth]{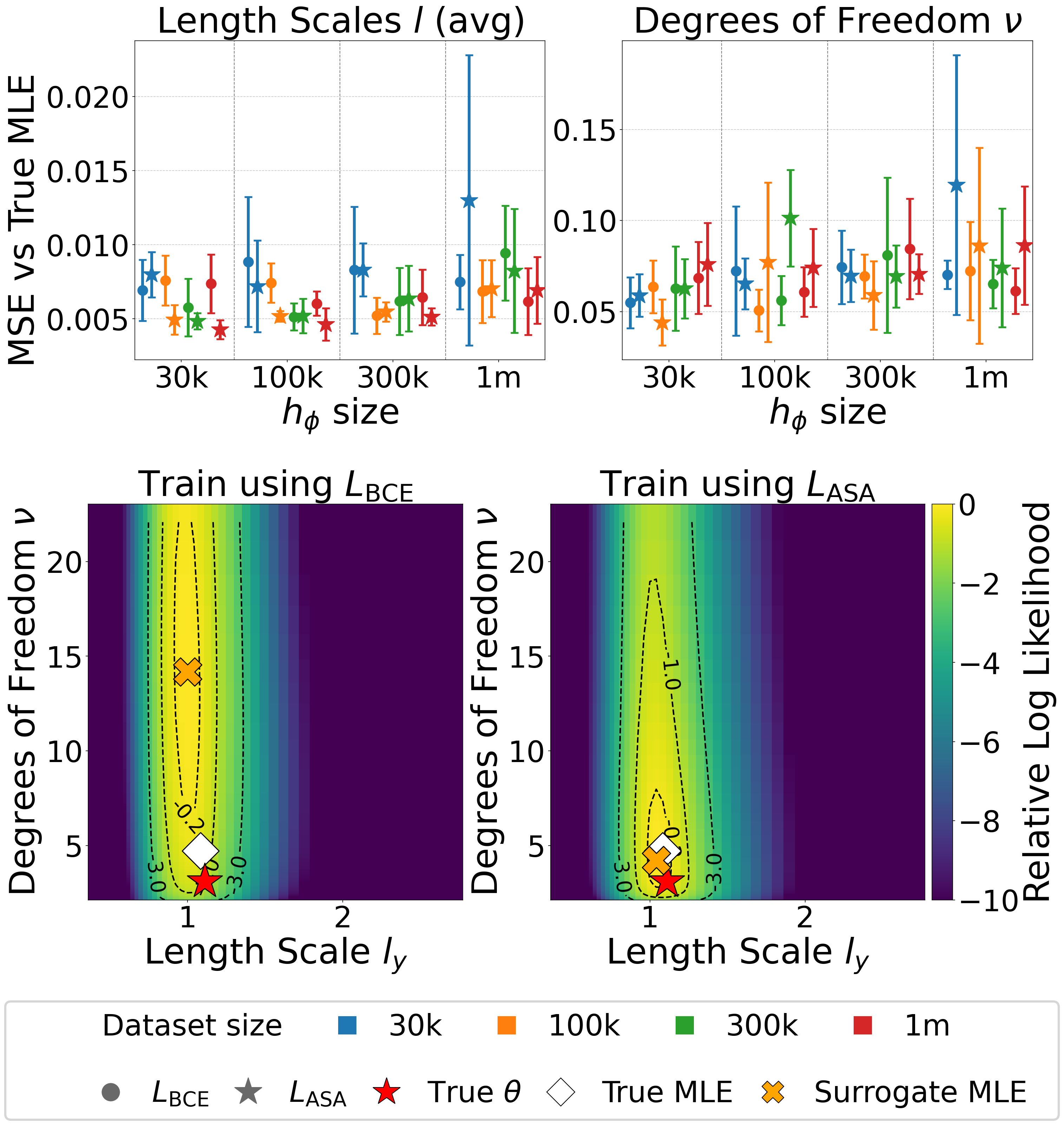}
    \caption{\textbf{Student-$t$ spatial process maximum likelihood estimator (MLE) metrics and example likelihood surfaces for different surrogate settings:} Figure is formatted identically to Figure~\ref{fig:sis_3}}
    \label{fig:stp_3}
\end{figure}

\clearpage
\subsection{Inference time}
\label{app:inference_time}

Table~\ref{tab:timing} compares inference time between direct evaluation of the ground truth likelihood to the best surrogate architecture for each case study. Times are averages $\pm2$ standard errors over the E-test set. Times are shown per fixed sized batch of evaluation points to control for effects of GPU batching.

Due to parameterization differences, the likelihood term $\mathbf{x}^\top \Sigma^{-1} \mathbf{x}$ requires inversion for every point in the parameter space for the Gaussian process, but only for unique values of length scales $(l_x, l_y)$ for the Student-$t$ process. The degrees of freedom $\nu$ do not affect $\Sigma$. Thus, we observe a large difference in per-batch ground truth inference times between the spatial case studies.

\begin{table}[ht]
\centering
\caption{Inference time comparisons between ground truth and surrogate likelihoods}
\label{tab:timing}
\begin{tabular}{llr}
\toprule
Setting & Method & Mean (ms) $\pm$ SEM \\
\midrule
SIS CT-HMM & Surrogate (size 100k) & $3.836 \pm 0.049$ \\
 & Ground truth & $1738.709 \pm 0.023$ \\
\midrule
Gaussian process & Surrogate (size 100k) & $24.687 \pm 0.002$ \\
 & Ground truth & $6587.082 \pm 0.097$ \\
\midrule
Student-$t$ process & Surrogate (size 30k) & $13.519 \pm 0.002$ \\
 & Ground truth & $106.704 \pm 0.011$ \\
\bottomrule
\end{tabular}
\end{table}

\subsection{Further timing studies of the different components of \ours}
\label{app:cost_breakdown}

We expand the results given in Figure~\ref{fig:overall_timing}, showing the contribution of each component of \ours to training cost. Tables~\ref{tab:sis_timing_breakdown}, \ref{tab:gp_timing_breakdown}, and \ref{tab:stp_timing_breakdown} show per-batch training time for different variations of \ours for each case study. Values shown are averages and standard errors over $\sim$ 9000 minibatches.
\begin{table}[ht]
\centering
\caption{\ours timing breakdown for SIS CT-HMM case study}
\begin{tabular}{llc}
\toprule
$\alpha$ update procedure & $\nabla_\theta h_\phi(x, \theta)$ calculation method & ms/batch (mean $\pm2$ SEM) \\
\midrule
Baseline NRE-A (no score loss) & -- & $6.15 \pm 0.02$ \\
Fixed $\alpha$ & Exact autograd & $11.15 \pm 0.04$ \\
Fixed $\alpha$ & Finite difference & $8.05 \pm 0.04$ \\
Adaptive $\alpha$, updated every batch & Exact autograd & $13.75 \pm 0.06$ \\
Adaptive $\alpha$, updated every batch & Finite difference & $11.11 \pm 0.04$ \\
Adaptive $\alpha$, updated every 64 batches & Exact autograd & $10.77 \pm 0.06$ \\
\textbf{Adaptive $\alpha$, updated every 64 batches} & Finite difference & $7.90 \pm 0.04$ \\
\bottomrule
\end{tabular}
\label{tab:sis_timing_breakdown}
\end{table}

\begin{table}[ht]
\centering
\caption{\ours timing breakdown for anisotropic Gaussian spatial process case study}
\begin{tabular}{llc}
\toprule
$\alpha$ update procedure & $\nabla_\theta h_\phi(x, \theta)$ calculation method & ms/batch (mean $\pm2$ SEM) \\
\midrule
Baseline NRE-A (no score loss) & -- & $8.51 \pm 0.06$ \\
Fixed $\alpha$ & Exact autograd & $13.46 \pm 0.06$ \\
Fixed $\alpha$ & Finite difference & $10.14 \pm 0.08$ \\
Adaptive $\alpha$, updated every batch & Exact autograd & $15.24 \pm 0.04$ \\
Adaptive $\alpha$, updated every batch & Finite difference & $13.67 \pm 0.06$ \\
Adaptive $\alpha$, updated every 64 batches & Exact autograd & $12.23 \pm 0.04$ \\
\textbf{Adaptive $\alpha$, updated every 64 batches} & Finite difference & $9.98 \pm 0.06$ \\
\bottomrule
\end{tabular}
\label{tab:gp_timing_breakdown}
\end{table}

\begin{table}[ht]
\centering
\caption{\ours timing breakdown for anisotropic Student-$t$ spatial process case study}
\begin{tabular}{llc}
\toprule
$\alpha$ update procedure & $\nabla_\theta h_\phi(x, \theta)$ calculation method & ms/batch (mean $\pm2$ SEM) \\
\midrule
Baseline NRE-A (no score loss) & -- & $5.09 \pm 0.04$ \\
Fixed $\alpha$ & Exact autograd & $8.80 \pm 0.04$ \\
Fixed $\alpha$ & Finite difference & $6.28 \pm 0.02$ \\
Adaptive $\alpha$, updated every batch & Exact autograd & $11.91 \pm 0.04$ \\
Adaptive $\alpha$, updated every batch & Finite difference & $9.77 \pm 0.04$ \\
Adaptive $\alpha$, updated every 64 batches & Exact autograd & $8.83 \pm 0.04$ \\
\textbf{Adaptive $\alpha$, updated every 64 batches} & Finite difference & $6.76 \pm 0.04$ \\
\bottomrule
\end{tabular}
\label{tab:stp_timing_breakdown}
\end{table}

\end{document}